\PassOptionsToClass{natbib=false}{acmart}
\documentclass[dvipsnames,sigconf,nonacm]{acmart}

\usepackage{tikz}
\usepackage{subcaption}
\usepackage[ruled,vlined]{algorithm2e}
\usepackage{placeins}

\AtBeginDocument{%
  }

\setcopyright{none}
\RequirePackage[
  backend=biber,
  datamodel=acmdatamodel,
  style=acmnumeric,
  ]{biblatex}

\begin{document}

\title{Continual Evolution Strategies in Control Tasks}
\titlenote{Accepted for publication in the GECCO 2026 Companion Proceedings.}

\author{Nicola Pitzalis}
\orcid{0009-0007-1169-9506}
\email{pitzalis93@gmail.com}
\affiliation{%
  \institution{University of Pisa}
  \city{Pisa}
  \country{Italy}
}

\author{Eleni Nisioti}
\orcid{0000-0001-7170-7108}
\email{enis@itu.dk}
\affiliation{%
  \institution{IT University of Copenhagen}
  \city{Copenhagen}
  \country{Denmark}
}

\author{Antonio Carta}
\orcid{0000-0002-0003-2323}
\email{antonio.carta@unipi.it}
\affiliation{%
  \institution{University of Pisa}
  \city{Pisa}
  \country{Italy}
}

\author{Davide Bacciu}
\orcid{0000-0001-5213-2468}
\email{davide.bacciu@unipi.it}
\affiliation{%
  \institution{University of Pisa}
  \city{Pisa}
  \country{Italy}
}

\author{Andrea Cossu}
\orcid{0000-0002-4874-8830}
\email{andrea.cossu@unipi.it}
\affiliation{%
  \institution{University of Pisa}
  \city{Pisa}
  \country{Italy}
}

\renewcommand{\shortauthors}{Pitzalis et al.}

\begin{abstract}
We study Evolution Strategies (ES) for continual control, where agents must adapt to changing tasks without forgetting previous ones. On sequential MuJoCo locomotion tasks, naive ES suffers from severe catastrophic forgetting. Replay substantially improves retention and can induce positive transfer, while larger replay budgets reduce plasticity. Overall, these results show that ES can support continual adaptation in control and that replay is an effective mechanism for mitigating forgetting.
\end{abstract}


\maketitle
\thispagestyle{plain}
\pagestyle{plain}
\section{Introduction}

In the current era of learning-based agents, controllers are typically optimized under the assumption that the environment dynamics and the agent morphology remain stationary over time \cite{xie2021a}. When these assumptions are violated, performance can degrade sharply, producing brittle agents with limited ability to adapt to changing conditions \cite{khetarpal2022,sim2real}. In realistic settings, however, it is unlikely that all operating conditions can be specified in advance. Instead, we should aim to build controllers that can continue adapting when the task changes, without sacrificing previously acquired behaviors. This is one of the central objectives of continual learning \cite{parisi2019, lesort2020}.

\begin{figure}[t]
    \centering
    \includegraphics[width=0.8\linewidth]{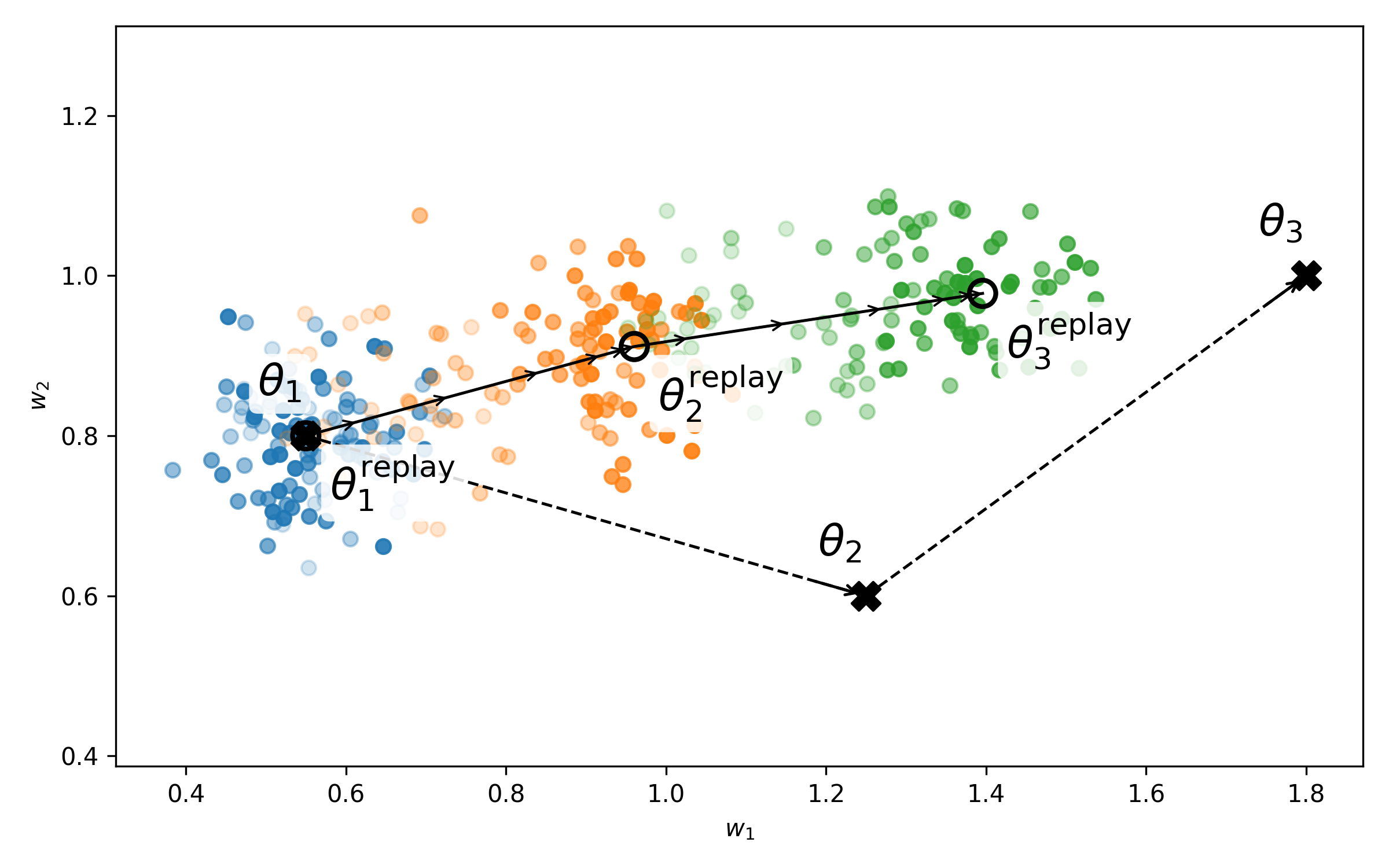}
    \caption{Conceptual illustration of continual learning with Evolution Strategies in parameter space $(w_1,w_2)$. Crosses denote single-task optima $\theta_i$, hollow circles replay-constrained solutions $\theta_i^{\mathrm{replay}}$, and the solid path the optimization trajectory under replay. Replay steers the search toward compromise regions that reduce forgetting.}
    \Description{Conceptual illustration of continual learning with Evolution Strategies in parameter space $(w_1,w_2)$. Crosses denote single-task optima $\theta_i$, hollow circles replay-constrained solutions $\theta_i^{\mathrm{replay}}$, and the solid path the optimization trajectory under replay. Replay steers the search toward compromise regions that reduce forgetting.}
    \label{fig:figure1}
\end{figure}

Continual learning has been studied extensively for gradient-based methods, especially in supervised and reinforcement learning \cite{masana2023,khetarpal2022}. By contrast, much less is known about population-based black-box optimization in this regime. Evolution Strategies (ES) optimize a distribution of candidate solutions rather than a single model, and have proved effective in continuous control \cite{salimans2017es,wierstra2014}. However, their behavior under strictly sequential task exposure remains largely unexplored.

In this work, we study ES in a continual control setting based on sequential MuJoCo locomotion tasks \cite{todorov2012mujoco}. We show that naive sequential ES adapts to new tasks but suffers from catastrophic forgetting \cite{french1999}. We then show that replay can be incorporated naturally into ES, substantially improving retention \cite{rolnick2019experience,hayes2021} and, in some cases, enabling positive transfer \cite{lopezpaz2017gem}. Taken together, these results provide, to our knowledge, the first empirical characterization of ES in a continual control setting and position ES as a viable optimization framework for non-stationary task streams.

\section{Related Work} \label{sec:background}

Continual learning studies how an agent can adapt to a sequence of tasks without overwriting previously acquired knowledge \cite{parisi2019,lesort2020}. In neural networks, sequential training typically leads to catastrophic forgetting \cite{french1999}. Replay (interleaving training on new experiences with training on previous ones) is one of the most common mitigation strategies \cite{hayes2021,rolnick2019experience}. Of particular importance is the task-agnostic setting of continual learning \cite{delange2021}, where the agent has no knowledge about which task is currently facing. In this challenging scenario, the same policy must solve all tasks without an explicit task label.

Evolutionary approaches for continual adaptation have been rarely explored. CEM-ACER \cite{tang2021cemacer} and ES-TD3 Buffers \cite{callaghan2023estd3buffers} combine evolutionary search with replay-based actor-critic updates, while \cite{lu2025ecl} studies evolutionary continual learning in supervised classification. Our work specifically focuses on ES-based control and quantifies forgetting, transfer, and adaptation: three of the most important dimensions of continual learning.

\section{Continual Evolution Strategies}

We study Evolution Strategies (ES) in a continual learning setting for continuous control. Let $\mathcal{T}=(T_1,\dots,T_n)$ denote an ordered sequence of tasks, and let $\theta \in \mathbb{R}^d$ be the parameter vector of a single policy trained sequentially across the stream. Training proceeds task by task, without resetting the parameters between tasks, yielding a parameter trajectory
\[
\theta^{(0)} \rightarrow \theta^{(1)} \rightarrow \cdots \rightarrow \theta^{(n)}.
\]
When optimizing task $T_i$, the policy is updated for a fixed number of ES generations, after which training continues on $T_{i+1}$. This protocol allows us to evaluate adaptation to new tasks, retention of previously acquired behaviors, and transfer across tasks.

We consider three MuJoCo continuous-control environments \cite{todorov2012mujoco}: \textit{Hopper-v5}, \textit{Walker2d-v5}, and \textit{Swimmer-v5}. These tasks differ in morphology and control structure, inducing varying degrees of similarity and conflict under sequential training. Policies are parameterized as shallow feedforward neural networks with task-specific input projections to account for heterogeneous observation spaces, followed by a shared hidden layer and a shared output head. This corresponds to the task-agnostic continual learning setting considered in this work.

Optimization follows a standard parallel ES procedure \cite{salimans2017es,wierstra2014}. At generation $t$, perturbations $\varepsilon_i \sim \mathcal{N}(0,I)$ are sampled to produce candidate parameters $\theta_t+\sigma \varepsilon_i$, where $\sigma$ is the noise scale. Let $F_i$ denote the return obtained by the perturbed policy. After rank-normalization of fitness values, parameters are updated according to
\[
\theta_{t+1}
=
\theta_t
+
\frac{\eta}{\lambda \sigma}
\sum_{i=1}^{\lambda} F_i \varepsilon_i,
\]
where $\lambda$ is the population size and $\eta$ is the learning rate. Training is performed for a fixed budget of generations on each task.

To mitigate forgetting, replay is implemented by adding extra fitness evaluations on previously encountered tasks during training on the current task. Replay samples are treated exactly as standard ES samples: all fitness values are rank-normalized jointly and contribute to the update with the same weight. Replay therefore changes only the set of sampled returns used to estimate the update, biasing the search toward solutions that retain performance across tasks.

Performance is measured through episodic return. Since reward scales differ across environments, we report the normalized reward on task $T_i$ at evaluation step $k$ as
\[
\hat{R}^{(k)}_{T_i}=\frac{R^{(k)}_{T_i}}{R^{*}_{\mathrm{base}_i}},
\]
where $R^{(k)}_{T_i}$ is the observed return and $R^{*}_{\mathrm{base}_i}$ is the maximum return obtained by training the same policy from scratch on task $T_i$ alone.

To quantify retention, we use backward transfer (BWT) \cite{lopezpaz2017gem}. For a task sequence $\mathcal{T}=(T_1,\dots,T_n)$, the average BWT is
\[
\mathrm{avg\ BWT}
=
\frac{1}{|\mathcal{T}_{\le n-1}|}
\sum_{T_i \in \mathcal{T}_{\le n-1}}
\mathrm{BWT}(T_i \leftarrow T_n,\mathrm{last}),
\]
where
\[
\mathrm{BWT}(T_i \leftarrow T_t,k)
=
\hat{R}^{(k)}_{T_i \mid \mathcal{T}_{\le t}}
-
\hat{R}^{*}_{T_i \mid \mathcal{T}_{\le i}}.
\]
Here, $\hat{R}^{(k)}_{T_i \mid \mathcal{T}_{\le t}}$ denotes the normalized return on task $T_i$ after training up to task $T_t$, and $\hat{R}^{*}_{T_i \mid \mathcal{T}_{\le i}}$ is the best normalized performance reached immediately after learning $T_i$. Negative BWT indicates forgetting, while positive values indicate beneficial backward transfer.

To quantify transfer to future tasks, we also report forward transfer (FWT) \cite{lopezpaz2017gem}. Its average value over a task sequence is
\[
\mathrm{avg\ FWT}
=
\frac{1}{|\mathcal{T}_{\ge 2}|}
\sum_{T_i \in \mathcal{T}_{\ge 2}}
\mathrm{FWT}(T_i \rightarrow T_{i-1},\mathrm{last}),
\]
with
\[
\mathrm{FWT}(T_i \rightarrow T_t,k)
=
\hat{R}^{(k)}_{T_i \mid \mathcal{T}_{\le t}}
-
\hat{R}_{T_i \mid \mathrm{rand}},
\qquad t<i,
\]
where $\hat{R}_{T_i \mid \mathrm{rand}}$ is the normalized return of a randomly initialized policy on task $T_i$. Positive FWT indicates that earlier tasks facilitate subsequent learning.

To quantify how previously learned tasks affect the acquisition of the current one, we define interference at acquisition for a task sequence $\mathcal{T}$ and position $\rho$ as
\[
\mathrm{Int}(\mathcal{T},\rho)=1-\hat{R}^{*}_{T_\rho \mid \mathcal{T}_{\le \rho}},
\]
where $\hat{R}^{*}_{T_\rho \mid \mathcal{T}_{\le \rho}}$ is the best normalized return achieved on task $T_\rho$ during its learning phase. Thus, $\mathrm{Int}=0$ indicates single-task performance, $\mathrm{Int}>0$ inhibitory interference, and $\mathrm{Int}<0$ facilitation.

\section{Experiments}
\label{sec:experiments}
\begin{figure}[t]
    \centering
    \includegraphics[width=0.7\linewidth]{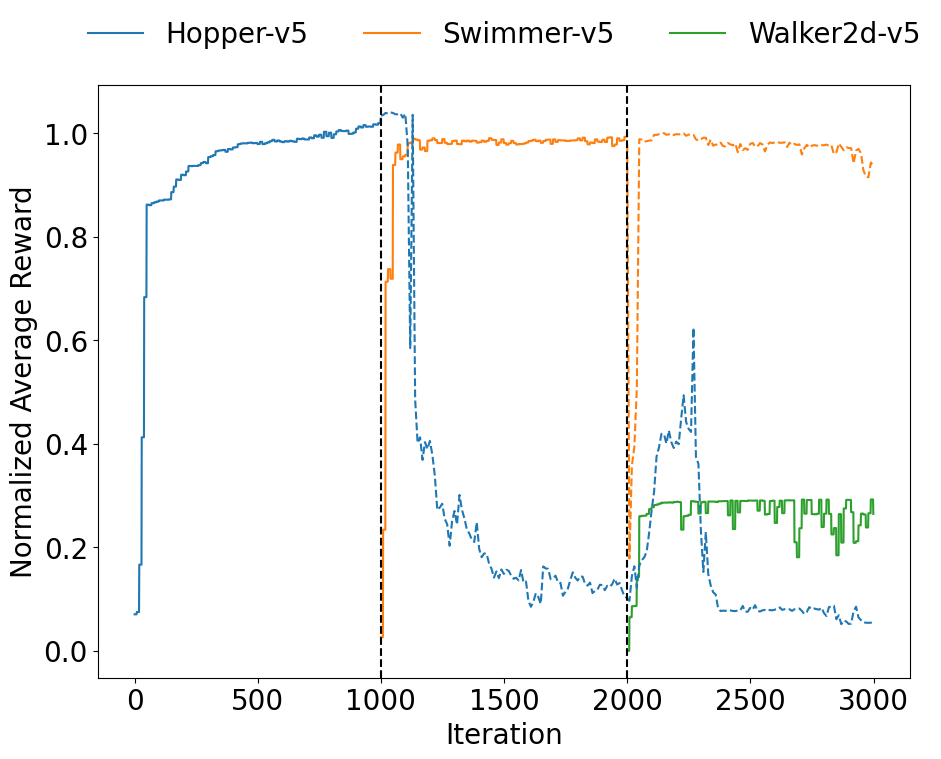}
    \caption{Normalized reward curves for the task-agnostic CL setting. Normalization is computed with respect to the maximum reward achieved in the corresponding single-task runs. The vertical dashed lines mark task transitions.}
    \Description{Normalized reward curves for the task-agnostic CL setting. Normalization is computed with respect to the maximum reward achieved in the corresponding single-task runs. The vertical dashed lines mark task transitions.}
    \label{fig:sequential_hsw}
\end{figure}

For each experiment, we report normalized reward curves over training together with the corresponding absolute maxima and minima (Supplementary material, Sec. F). Rewards are normalized by the maximum return achieved in the corresponding single-task baseline, so all plots are expressed relative to the best single-task performance attainable on each environment. In the curves, the first task is shown as a solid line, corresponding to standard single-task training, whereas subsequent tasks are shown as dashed lines, corresponding to sequential continual training.

To ensure comparability across tasks, performance is measured on the unperturbed policy every 10 iterations. Each evaluation averages 10 rollouts over different environment seeds. Unless otherwise stated, experiments use a fixed task order; additional task permutations are considered separately to assess order effects. Hyperparameters were chosen via model selection on a single task permutation and then kept fixed for all remaining runs. The population size $\lambda$ is $768\pm384$, the standard deviation $\sigma$ of noise is $0.1$, the learning rate $\eta$ is $0.05$. We used $1000$ ES generations per task and a replay buffer of $12, 192$ and $288$.

\subsection{Task-Agnostic Continual Learning}

We consider a task-agnostic continual setting in which the learner is trained on the first task with standard ES and then updated sequentially on subsequent tasks without resetting the parameters. This enforces strict parameter sharing across tasks: only the input layer is task-specific, whereas the hidden and output layers are shared.

Figure~\ref{fig:sequential_hsw} shows that sequential ES adapts to each newly introduced task, but does not preserve previously acquired behaviors. Performance on \textit{Hopper} drops sharply after training on later tasks, and \textit{Swimmer} also degrades after the transition to \textit{Walker2d}. This behavior is a clear instance of catastrophic forgetting under sequential ES.

\subsection{Replay to Mitigate Forgetting}

We next investigate replay as a mechanism to reduce forgetting under sequential ES. We vary the replay budget in order to assess whether small amounts of replay are sufficient to stabilize learning and whether larger budgets eventually saturate or hinder adaptation.

\begin{figure}[t]
    \centering
    \includegraphics[width=0.7\linewidth]{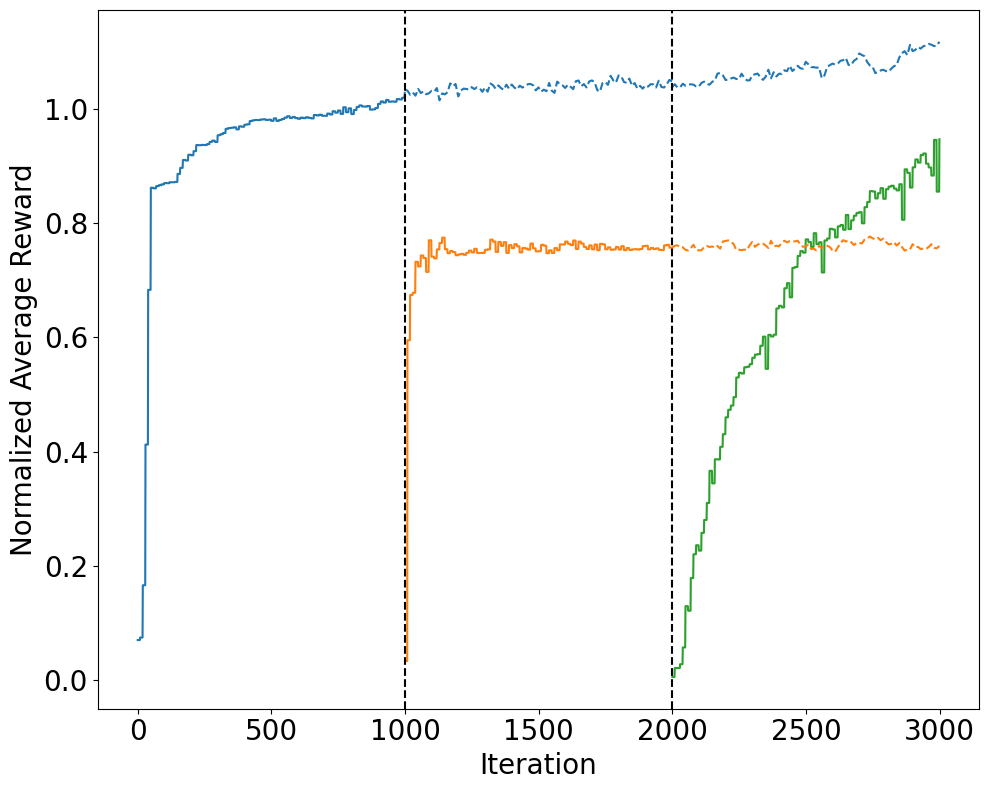}
    
    \caption{Normalized reward curves for the task-agnostic CL setting with replay (+192 steps, $\sim$25\%). Normalization is computed with respect to the maximum reward achieved in the corresponding single-task runs. The vertical dashed lines mark task transitions.}
    \Description{Normalized reward curves for the task-agnostic CL setting with replay (+192 steps, about 25 percent). Normalization is computed with respect to the maximum reward achieved in the corresponding single-task runs. The vertical dashed lines mark task transitions.}
    \label{fig:sequential_hsw_replay}
    \end{figure}

Replay substantially reduces the abrupt post-switch performance collapse observed in the no-replay setting (Figure~\ref{fig:sequential_hsw_replay}). Previously learned tasks remain considerably more stable after later tasks are introduced, while newly encountered tasks can still be acquired. In this sense, replay acts as an effective regularizer for continual ES, biasing the search toward parameter regions that preserve performance across tasks.

Increasing replay also reveals a stability--plasticity trade-off. With small replay budgets, adaptation to the current task remains easy but retention is only partial. Intermediate replay yields the best compromise, improving retention without strongly impairing learning on later tasks. At the largest budget considered, stability dominates: earlier tasks are preserved more effectively, but plasticity is reduced and later tasks are fitted less efficiently. Overall, these results show that replay is a simple and effective mechanism for mitigating catastrophic forgetting in continual evolutionary optimization.

\begin{figure}[t]
    \centering
    \includegraphics[width=0.7\linewidth]{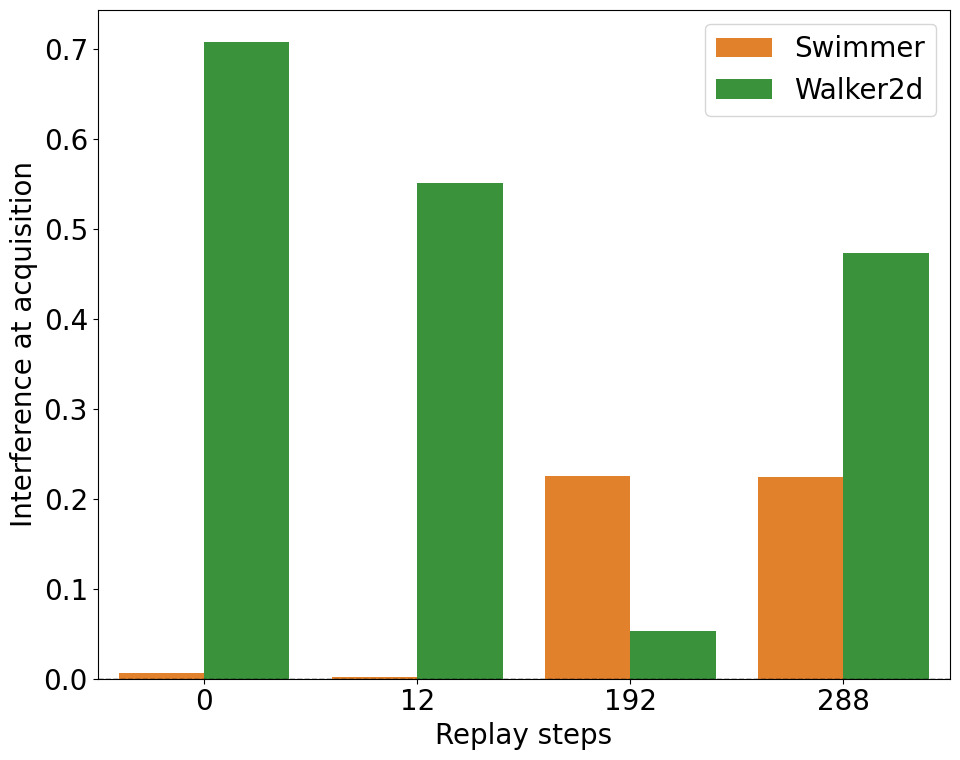}
    \caption{Interference at acquisition for the task sequence $\mathcal{T}=(\textit{Hopper},\,\textit{Swimmer},\,\textit{Walker2d})$ under different replay budgets. Bars report $\mathrm{Int}(\mathcal{T},\rho)$ at the acquisition step of each task. Lower values indicate reduced interference.}
    \Description{Interference at acquisition for the task sequence Hopper, Swimmer, Walker2d under different replay budgets. Bars report interference values at the acquisition step of each task. Lower values indicate reduced interference during learning.}
    \label{fig:interference_replay}
\end{figure}
To further characterize this trade-off, Figure~\ref{fig:interference_replay} reports interference at acquisition under different replay budgets. With little or no replay, later tasks experience stronger interference during learning. Intermediate replay reduces this effect, yielding the best balance between retention and adaptability. At the largest replay budget, interference increases again, indicating that overly strong stabilization can hinder the acquisition of new tasks.

\subsection{Knowledge Transfer Across Tasks}

We next study whether replay-based sequential ES also induces knowledge transfer across tasks. To this end, we measure backward transfer (BWT) and forward transfer (FWT) over all permutations of the task sequence.

\begin{figure}[t]
     \centering
    \includegraphics[width=0.48\linewidth]{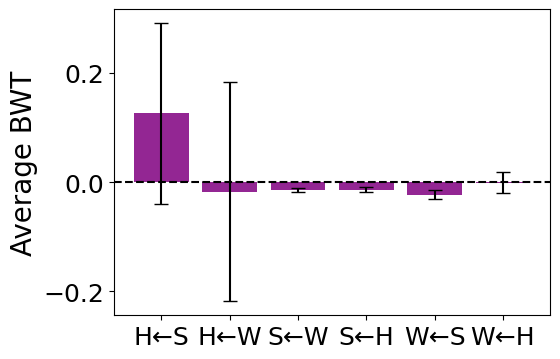}
    \includegraphics[width=0.48\linewidth]{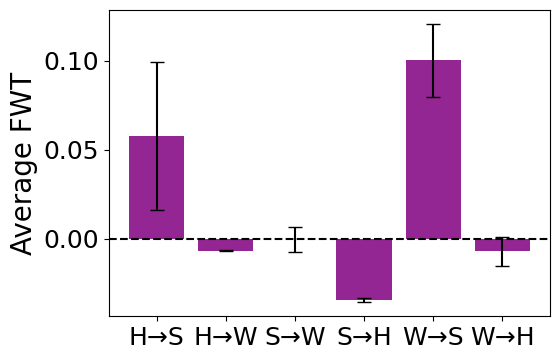}

    \caption{Pairwise transfer averaged over sequence contexts. Top: backward transfer (BWT), where $A \leftarrow B$ measures the effect on task $A$ after learning task $B$. Bottom: forward transfer (FWT), where $A \rightarrow B$ measures the effect of learning task $A$ before task $B$. Positive values indicate beneficial transfer; negative values indicate forgetting for BWT and harmful transfer for FWT. Error bars denote variability across contexts.}
    \Description{Pairwise transfer averaged over sequence contexts. Top: backward transfer, where A leftarrow B measures the effect on task A after learning task B. Bottom: forward transfer, where A rightarrow B measures the effect of learning task A before task B. Positive values indicate beneficial transfer; negative values indicate forgetting for backward transfer and harmful transfer for forward transfer. Error bars denote variability across contexts.}
    \label{fig:bwt_fwt}
\end{figure}

Figure~\ref{fig:bwt_fwt} shows that transfer is limited and strongly asymmetric. Backward transfer is generally close to zero, indicating that replay keeps forgetting under control across most task orderings. A notable exception is the pair \textit{Hopper}$\leftarrow$\textit{Swimmer}, which exhibits positive BWT, although with high variability across curricula.
Forward transfer is more selective. Positive FWT is observed mainly toward \textit{Swimmer}, for both \textit{Hopper}$\rightarrow$\textit{Swimmer} and \\ \textit{Walker2d}$\rightarrow$\textit{Swimmer}, suggesting that legged locomotion can provide useful structure for subsequent adaptation to this task. In contrast, transfer from \textit{Swimmer} to the legged agents is weak or negative, indicating lower compatibility between the corresponding parameter regions.
Overall, these results suggest that replay-based ES exhibits limited but non-negligible transfer, which emerges only for specific task orderings and compatible task pairs.

\section{Discussion and Future Work}

We studied Evolution Strategies in a continual control setting based on sequential locomotion tasks. Our results show that naive sequential ES suffers from catastrophic forgetting, but that replay substantially improves retention and can induce positive transfer. At the same time, increasing replay reveals a stability--plasticity trade-off: stronger replay better preserves earlier tasks, but can reduce adaptation to later ones.
These findings position ES as a viable optimization framework for continual learning in control. Although ES exhibit the same fundamental interference issues as gradient-based methods, replay integrates naturally into the evolutionary setting, requiring neither gradient information nor architectural changes. From this perspective, replay can be understood as biasing population-based search toward parameter regions that perform well across multiple tasks rather than over-specializing to the current one.

Transfer effects were selective rather than uniform across task orderings. Forward transfer was generally limited and asymmetric, while backward transfer remained close to zero under replay, indicating stable retention in most cases. Beneficial transfer emerged only for specific task pairs, suggesting that it depends on the compatibility of their corresponding parameter optima.

A natural direction for future work is to combine ES with additional continual learning strategies. In particular, it would be interesting to investigate whether regularization-based methods, such as distillation \cite{li2016a} or importance-weighting approaches \cite{kirkpatrick2017}, can be adapted to black-box evolutionary optimization.

\begin{acks}
Work funded by the EU EIC project EMERGE (Grant No. 101070918) and European Union (ERC, GROW-AI, 101045094). Views and opinions expressed are however those of the authors only and do not necessarily reflect those of the European Union or the European Research Council.
\end{acks}

\printbibliography

\clearpage
\appendix
\section{Algorithm}
We report here the continual Evolution Strategies (ES) training procedure used in the main paper. Training proceeds sequentially over an ordered stream of tasks, without resetting the policy parameters between tasks. Replay is implemented by interleaving additional evaluations on previously encountered tasks during selected ES generations.

\begin{algorithm}[b]
\DontPrintSemicolon
\caption{Continual Evolution Strategies}
\label{alg:continual-es}

\KwIn{Tasks $\{T_1,\dots,T_n\}$,
population size $\lambda$, iterations per task $K$, learning rate $\eta$,
noise scale $\sigma$, Replay interval $R$ (if $R = 0$, no Replay),
Replay task $T_{\text{replay}}$ (a previously learned task)}
\KwOut{Final policy parameters $\theta$}

Initialize policy parameters $\theta$\;

\For{$i \leftarrow 1$ \KwTo $n$}{ \tcp*[r]{task loop}
  \For{$t \leftarrow 1$ \KwTo $K$}{
  
    \tcp{Select evaluation task for this iteration}
    \eIf{$R = 0$ \textbf{or} $t \bmod R \neq 0$}{
      $\tau \leftarrow T_i$ \tcp*{current-task iteration}
    }{
      $\tau \leftarrow T_{\text{replay}}$ \tcp*{replay iteration on a past task}
    }
    
    \For{$k \leftarrow 1$ \KwTo $\lambda$}{
      Sample perturbation $\varepsilon_k \sim \mathcal{N}(0, I)$\;
      $\theta_k \leftarrow \theta + \sigma \varepsilon_k$\;
      $F_k \leftarrow \text{Evaluate}(\theta_k, \tau)$ \tcp*{episodic return in task $\tau$}
    }
    
    Rank-normalize fitnesses $\{F_k\}_{k=1}^{\lambda}$\;
    $g \leftarrow \dfrac{1}{\lambda \sigma} \sum_{k=1}^{\lambda} F_k \varepsilon_k$\;
    $\theta \leftarrow \theta + \eta g$\;
  }
}

\Return $\theta$\;

\end{algorithm}

\section{The task-aware variant}
\label{sec:task-aware-variant}

\begin{figure}[H]
\centering

\begin{subfigure}{\linewidth}
\centering
\begin{tikzpicture}[font=\Large, scale=0.5, every node/.style={transform shape}]
  \node[draw, rounded corners, align=center, minimum width=3.8cm, minimum height=0.9cm] (InpA) at (-5.0,2.0) {Input layer (task A)};
  \node[draw, rounded corners, align=center, minimum width=3.8cm, minimum height=0.9cm] (InpB) at (-5.0,1.0) {Input layer (task B)};
  \node[draw, rounded corners, align=center, minimum width=3.8cm, minimum height=0.9cm] (InpC) at (-5.0,0.0) {Input layer (task C)};
  
  \node[anchor=east] at (-7.1,2.0) {obs A};
  \node[anchor=east] at (-7.1,1.0) {obs B};
  \node[anchor=east] at (-7.1,0.0) {obs C};

  \node[draw, rounded corners, align=center, minimum width=3.2cm, minimum height=1.8cm] (Hid) at (-0.5,1.0) {Shared hidden};

  \node[draw, rounded corners, align=center, minimum width=3.6cm, minimum height=0.9cm] (Head) at (4.0,1.0) {Shared output head\\(tanh actions)};

  \draw[->] (InpA.east) -- (Hid.west);
  \draw[->] (InpB.east) -- (Hid.west);
  \draw[->] (InpC.east) -- (Hid.west);
  \draw[->] (Hid.east) -- (Head.west);

  \node[anchor=west] at (6.0,1.0) {actions};
\end{tikzpicture}
\caption{Single shared output head with distinct task-specific input layers.}
\label{fig:single-shared-head}
\end{subfigure}

\begin{subfigure}{\linewidth}
\centering
\begin{tikzpicture}[font=\Large, scale=0.5, every node/.style={transform shape}]
  \node[draw, rounded corners, align=center, minimum width=3.8cm, minimum height=0.9cm] (InpA) at (-5.0,2.0) {Input layer (task A)};
  \node[draw, rounded corners, align=center, minimum width=3.8cm, minimum height=0.9cm] (InpB) at (-5.0,1.0) {Input layer (task B)};
  \node[draw, rounded corners, align=center, minimum width=3.8cm, minimum height=0.9cm] (InpC) at (-5.0,0.0) {Input layer (task C)};

  \node[anchor=east] at (-7.1,2.0) {obs A};
  \node[anchor=east] at (-7.1,1.0) {obs B};
  \node[anchor=east] at (-7.1,0.0) {obs C};

  \node[draw, rounded corners, align=center, minimum width=3.2cm, minimum height=1.8cm] (Hid) at (-0.5,1.0) {Shared hidden};

  \node[draw, rounded corners, align=center, minimum width=3.6cm, minimum height=0.9cm] (HeadA) at (4.0,2.0) {Output head (task A)};
  \node[draw, rounded corners, align=center, minimum width=3.6cm, minimum height=0.9cm] (HeadB) at (4.0,1.0) {Output head (task B)};
  \node[draw, rounded corners, align=center, minimum width=3.6cm, minimum height=0.9cm] (HeadC) at (4.0,0.0) {Output head (task C)};

  \draw[->] (InpA.east) -- (Hid.west);
  \draw[->] (InpB.east) -- (Hid.west);
  \draw[->] (InpC.east) -- (Hid.west);
  \draw[->] (Hid.east) -- (HeadA.west);
  \draw[->] (Hid.east) -- (HeadB.west);
  \draw[->] (Hid.east) -- (HeadC.west);

  \node[anchor=west] at (6.0,1.8) {actions A};
  \node[anchor=west] at (6.0,1.0) {actions B};
  \node[anchor=west] at (6.0,0.2) {actions C};
\end{tikzpicture}
\caption{Task-aware setting with distinct task-specific input and output layers.}
\label{fig:distinct-heads-both-ends}
\end{subfigure}

\caption{Comparison of task-agnostic (top) and task-aware (bottom) architectures.}
\Description{Comparison of task-agnostic (top) and task-aware (bottom) architectures.}
\label{fig:architectures}
\end{figure}
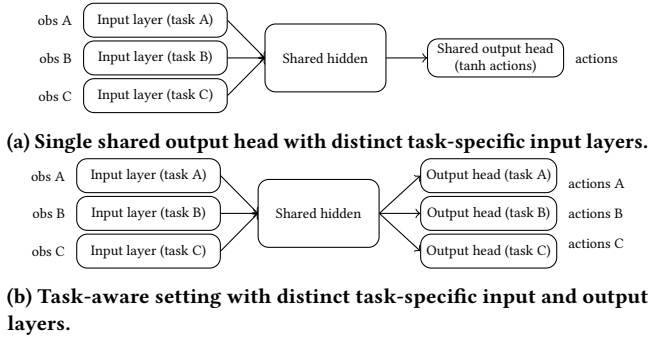

Policies are parameterized as shallow feedforward neural networks. Since the environments expose heterogeneous observation spaces, each task is first processed by a task-specific input projection layer. Formally, the policy for task $T_i$ is written as
\[
\pi_\theta^{(i)}(\cdot)
= h_{\theta_o^{(i)}}\!\big(f_{\theta_s}(g_{\theta_{in}^{(i)}}(\cdot))\big),
\]
where $g_{\theta_{in}^{(i)}}$ denotes the task-specific input projection, $f_{\theta_s}$ the shared hidden representation (a 64-unit ReLU layer), and $h_{\theta_o^{(i)}}$ the output head.

The parameter vector therefore decomposes as
\[
\theta=\{\theta_s,\theta_{in}^{(1:n)},\theta_o^{(1:n)}\},
\]
so that the hidden layer acts as the main shared representational bottleneck across tasks.

The task-agnostic and task-aware variants differ only in the definition of the output layer. In the task-agnostic setting, a single shared output head is used for all tasks, i.e.
\[
h_{\theta_o^{(i)}} \equiv h_{\theta_o} \qquad \text{for all } i.
\]
In the task-aware setting, instead, each task is assigned a distinct output head, so that the final action mapping becomes task-specific.

This architectural change is important in continual control. With a shared output head, the same output parameters must express action policies for morphologically different agents and environments. As a consequence, the output units acquire task-dependent semantics, making the output layer a direct source of interference and forgetting. In the task-aware variant, previously learned action mappings are isolated in separate heads, so updates on a new task do not overwrite the output parameters used by earlier tasks. Forgetting can still arise in the shared hidden representation, but interference at the action level is substantially reduced.

Figure~\ref{fig:architectures} compares the two variants. The top architecture corresponds to the task-agnostic case used in the main continual setting, while the bottom one corresponds to the task-aware multi-head variant discussed in the supplementary results.

\section{Baselines}
\label{sec:baselines}
\begin{figure}[tbp]
    \centering
    \includegraphics[width=\linewidth]{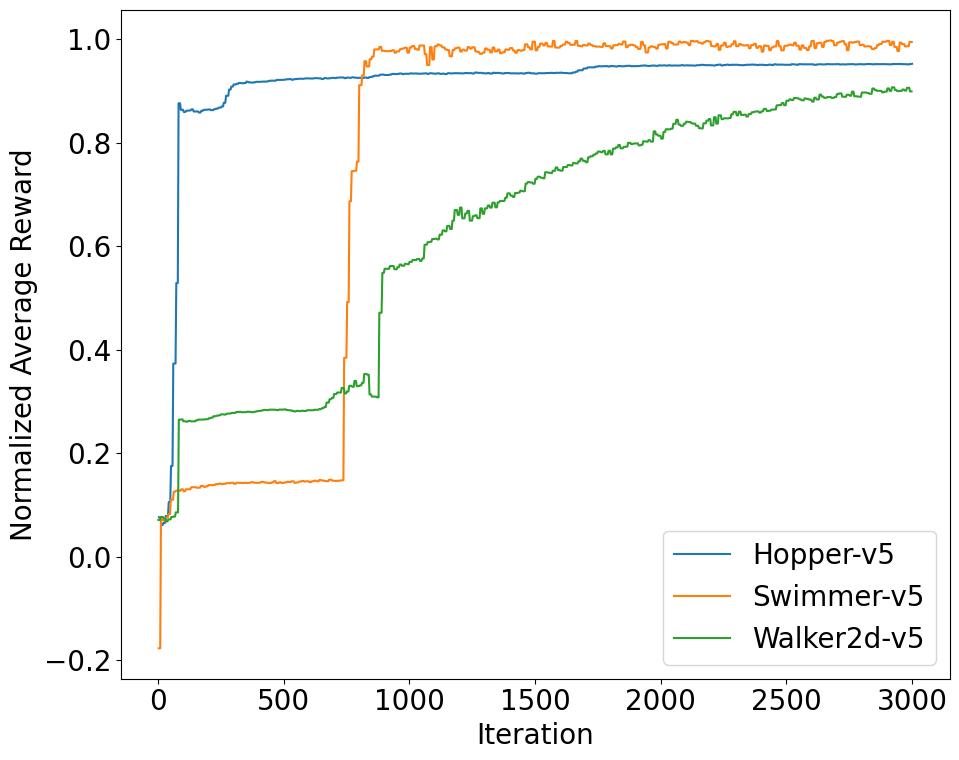}
    \caption{Baseline experiments for the three MuJoCo environments. Each curve shows the learning progress of a policy trained from scratch on a single task (\textit{Hopper-v5}, \textit{Swimmer-v5}, or \textit{Walker2d-v5}), reporting the normalized average reward over training iterations. Normalization is done with respect to the maximum reward achieved in the corresponding single-task run.}
    \Description{Baseline experiments for the three MuJoCo environments. Each curve shows the learning progress of a policy trained from scratch on a single task (\textit{Hopper-v5}, \textit{Swimmer-v5}, or \textit{Walker2d-v5}), reporting the normalized average reward over training iterations. Normalization is done with respect to the maximum reward achieved in the corresponding single-task run.}

    \label{fig:baselines}
\end{figure}

Here, we will report the experiments that act as a baseline for the continual setup, precisely we will show both the learning curves (Figure~\ref{fig:baselines}) of each of the three chosen tasks, as well as their un-normalized rewards (Table~\ref{tab:baseline-single-task}). These values are the ones used as denominators to normalize the continual learner's performance.

\begin{table}[b]
\centering
\begin{tabular}{l c}
\hline
\textbf{Task} & \textbf{Max. reward} \\
\hline
Walker2d-v5 & 4680.66 \\
Swimmer-v5  &  362.21 \\
Hopper-v5   & 3511.92 \\
\hline
\end{tabular}
\caption{Maximum evaluation returns for single-task baselines.}
\Description{Maximum evaluation returns for single-task baselines.}
\label{tab:baseline-single-task}
\end{table}

As a comparison measure, we also show the results obtained training the agent on a single dataset comprising of all the tasks together---thus in a non-continual setting. In this version, the agent runs on each environment alternating each of them with one another, step by step; the results are shown in Table~\ref{tab:baseline-mulit-task}.

\begin{table}[tbp]
\centering
\setlength{\tabcolsep}{4pt}
\begin{tabular}{lccc}
\hline
\textbf{Experiment} & \textbf{Hopper-v5} & \textbf{Swimmer-v5} & \textbf{Walker2d-v5} \\
\hline
Separate outputs & 1224.28 & 360.54 & 3719.17 \\
Shared output    & 1112.05 & 361.45 & 3396.36 \\
\hline
\end{tabular}
\caption{Maximum evaluation returns across tasks for the multi-task case.}
\Description{Maximum evaluation returns across tasks for the multi-task case.}
\label{tab:baseline-mulit-task}
\end{table}

\section{Task-aware continual control}
\label{sec:task-aware-results}

\begin{figure}[t]
    \centering
    \includegraphics[width=0.8\linewidth]{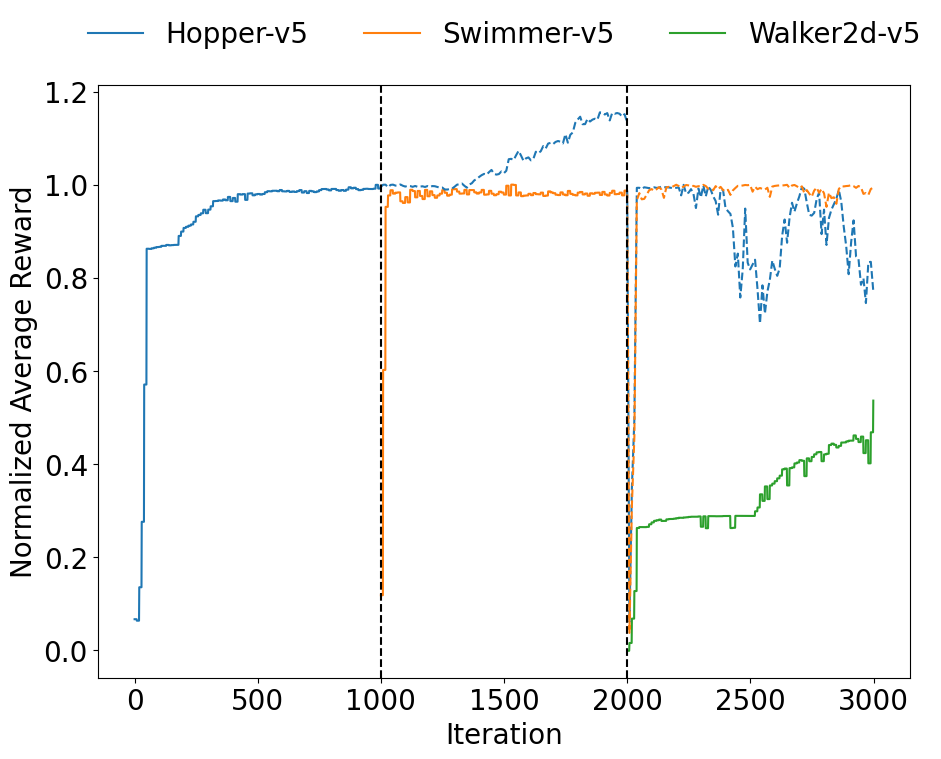}
    \caption{Normalized reward curves for the task-aware setting on the sequence \textit{Hopper} $\to$ \textit{Swimmer} $\to$ \textit{Walker2d}. Normalization is computed with respect to the maximum reward achieved in the corresponding single-task runs. Vertical dashed lines mark task transitions.}
    \Description{Normalized reward curves for the task-aware setting on the sequence Hopper to Swimmer to Walker2d. Normalization is computed with respect to the maximum reward achieved in the corresponding single-task runs. Vertical dashed lines mark task transitions.}
    \label{fig:sequential_hsw_distinct}
\end{figure}

\begin{figure}[t]
    \centering
    \includegraphics[width=0.7\linewidth]{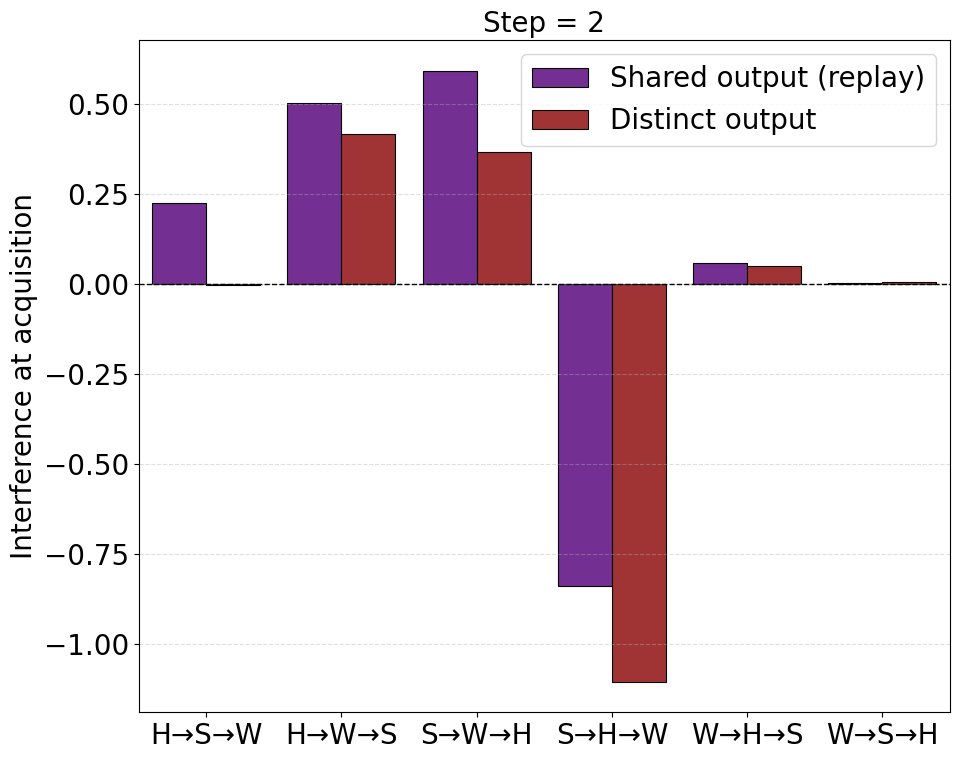}
    \includegraphics[width=0.7\linewidth]{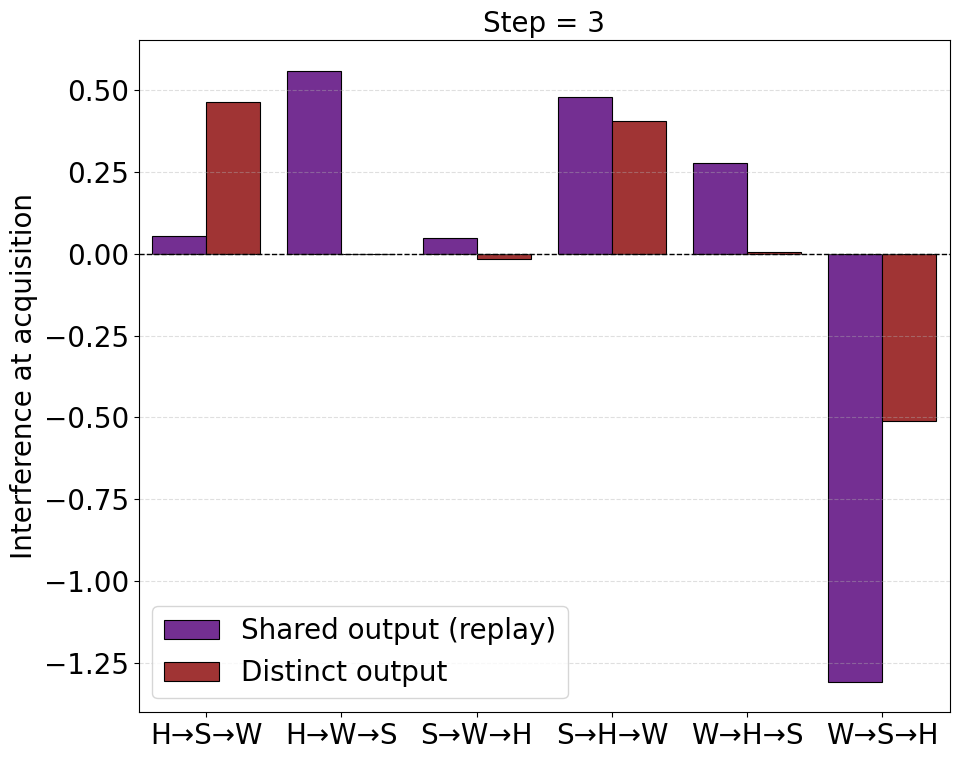}
    \caption{Comparison of interference at acquisition between the replay-based shared-output model (192 replay steps) and the task-aware distinct-output model, evaluated at the acquisition of the second task (top) and of the third task (bottom). Each bar corresponds to a different task sequence $\mathcal{T}$, with interference computed as $\mathrm{Int}(\mathcal{T}, \rho)$ at acquisition position $\rho$.}
    \Description{Comparison of interference at acquisition between the replay-based shared-output model and the task-aware distinct-output model, evaluated at the acquisition of the second task on top and of the third task on bottom.}
    \label{fig:interference_distinct_vs_replay}
\end{figure}

We now report the results obtained with the task-aware variant, in which policies share only the hidden representation, while both input and output layers remain task-specific. This corresponds to a task-incremental continual learning setting, where the task identity is available at inference time. In this regime, forgetting at the output level is largely removed by construction, so the observed differences in performance mainly reflect the extent to which the shared hidden representation can support multiple tasks without harmful interference.

Figure~\ref{fig:sequential_hsw_distinct} shows the learning dynamics for the sequence \textit{Hopper} $\to$ \textit{Swimmer} $\to$ \textit{Walker2d} under the distinct-output architecture. In contrast to the shared-output case, previously learned tasks remain largely stable after each task transition, showing that the task-aware variant strongly reduces catastrophic forgetting. In addition, some curves exhibit slight improvements after subsequent tasks are introduced, suggesting the presence of positive knowledge transfer through the shared hidden representation.

To better isolate the role of the output layer, we also compare interference at acquisition between the task-aware model and the replay-based shared-output model. If interference is high with a shared head but substantially reduced with task-specific heads, this indicates that a major source of conflict lies in the action mapping rather than in the shared representation. Figure~\ref{fig:interference_distinct_vs_replay} reports this comparison for the acquisition of the second task (top) and the third task (bottom), across all task permutations. In nearly all cases, interference is lower in the task-aware setting, confirming that the shared output layer is a major source of forgetting in these control environments.

\section{Replay budget study}
\label{sec:replay-budget}

\begin{figure}[t]
    \centering
    \includegraphics[width=0.8\linewidth]{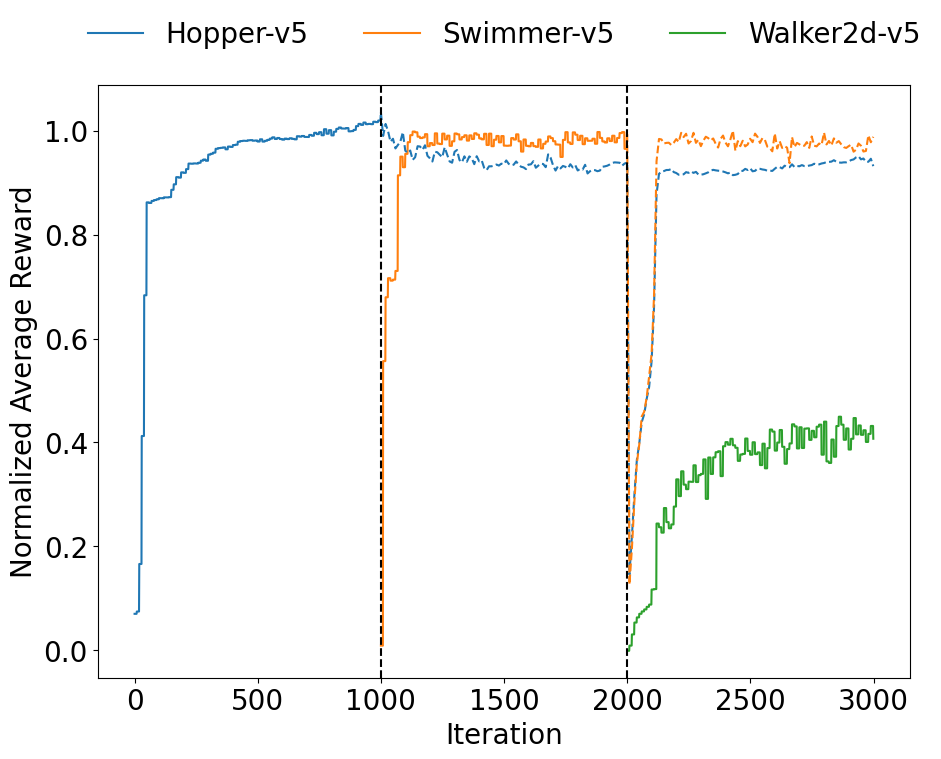}
    \includegraphics[width=0.8\linewidth]{replay_study_replay16.png}
    \includegraphics[width=0.8\linewidth]{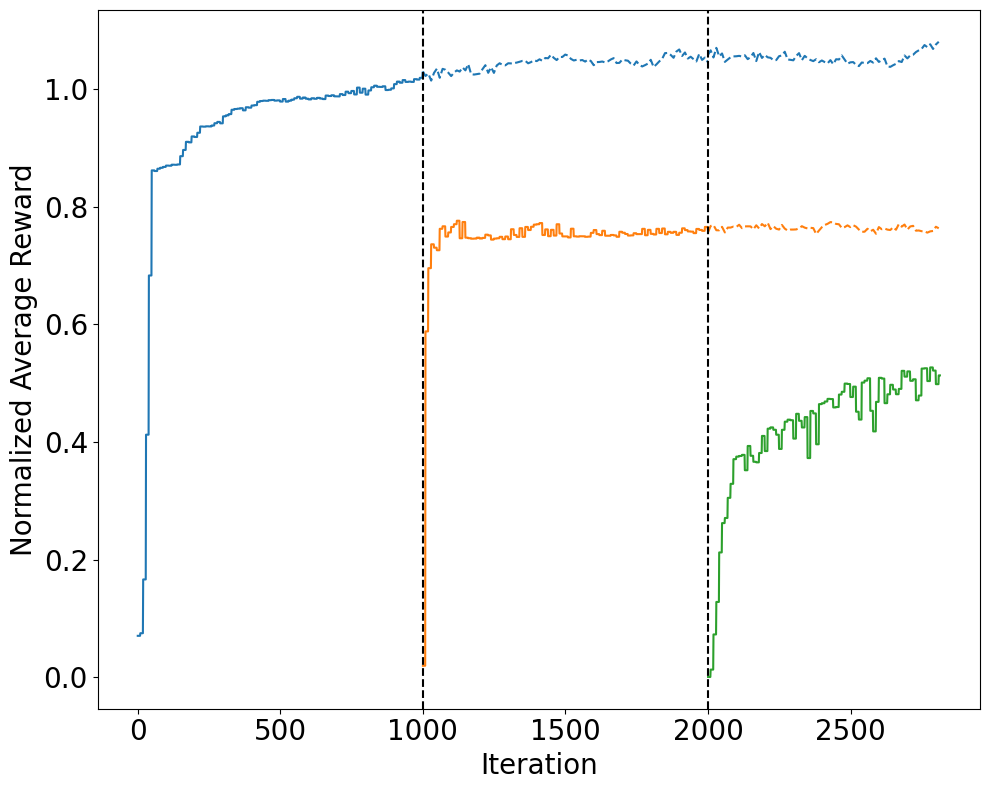}
    \caption{Normalized reward curves for the shared-output setting on the sequence \textit{Hopper} $\to$ \textit{Swimmer} $\to$ \textit{Walker2d} under different replay budgets. Replay configurations correspond to +12 steps ($\sim$1.5\%), +192 steps ($\sim$25\%), and +288 steps ($\sim$37.5\%). Normalization is computed with respect to the maximum reward achieved in the corresponding single-task runs. Vertical dashed lines mark task transitions.}
    \Description{Normalized reward curves for the shared-output setting on the sequence Hopper to Swimmer to Walker2d under different replay budgets. Replay configurations correspond to 12, 192, and 288 replay steps. Normalization is computed with respect to the maximum reward achieved in the corresponding single-task runs. Vertical dashed lines mark task transitions.}
    \label{fig:replay-budget-study}
\end{figure}

For completeness, we report the effect of varying the replay budget in the shared-output setting for the fixed sequence \textit{Hopper} $\to$ \textit{Swimmer} $\to$ \textit{Walker2d}. Figure~\ref{fig:replay-budget-study} compares three replay configurations, corresponding to 12, 192, and 288 replay perturbations. The figure illustrates the stability--plasticity trade-off induced by replay: small replay budgets preserve plasticity but provide weaker retention, intermediate replay yields the best balance between retention and adaptation, and large replay budgets further stabilize previously learned tasks at the cost of reduced plasticity on later ones.

To further characterize this effect, Figure~\ref{fig:interference_replay} reports interference at acquisition for the same task sequence under increasing replay budgets. Moderate replay substantially reduces interference on later tasks, whereas excessive replay makes learning more constrained and increases interference again. Together, these results show that replay acts as an effective stabilizing mechanism, but only up to a point, after which additional replay begins to hinder adaptation.

\section{All permutations}
\label{sec:all-permutations}
\begin{figure*}[p]
\centering
\setkeys{Gin}{height=0.23\textheight,keepaspectratio}

\begin{subfigure}{0.45\textwidth}
    \centering
    \includegraphics[width=\linewidth]{shared_hsw_first.png}
    \includegraphics[width=\linewidth]{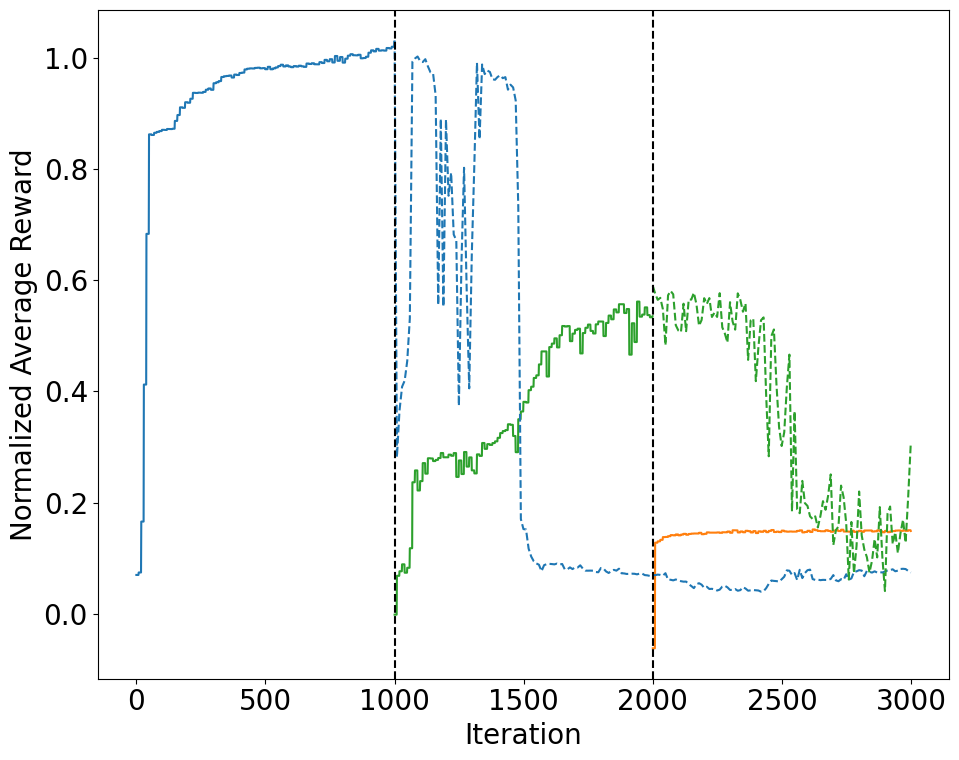}
    \includegraphics[width=\linewidth]{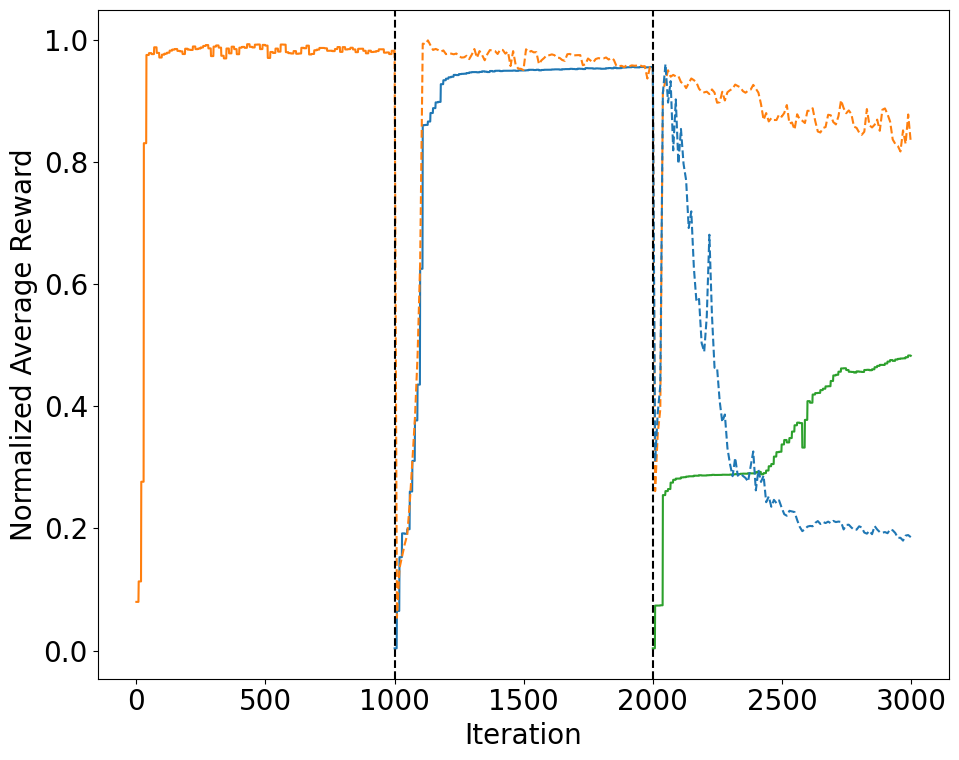}
\end{subfigure}\hfill
\begin{subfigure}{0.45\textwidth}
    \centering
    \includegraphics[width=\linewidth]{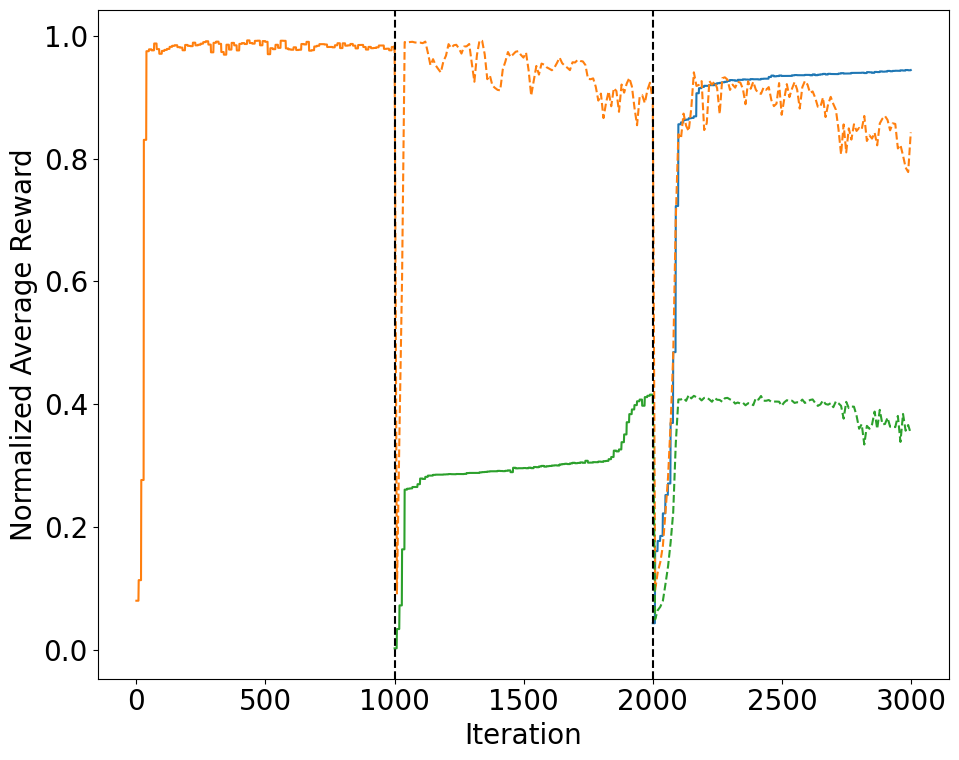}
    \includegraphics[width=\linewidth]{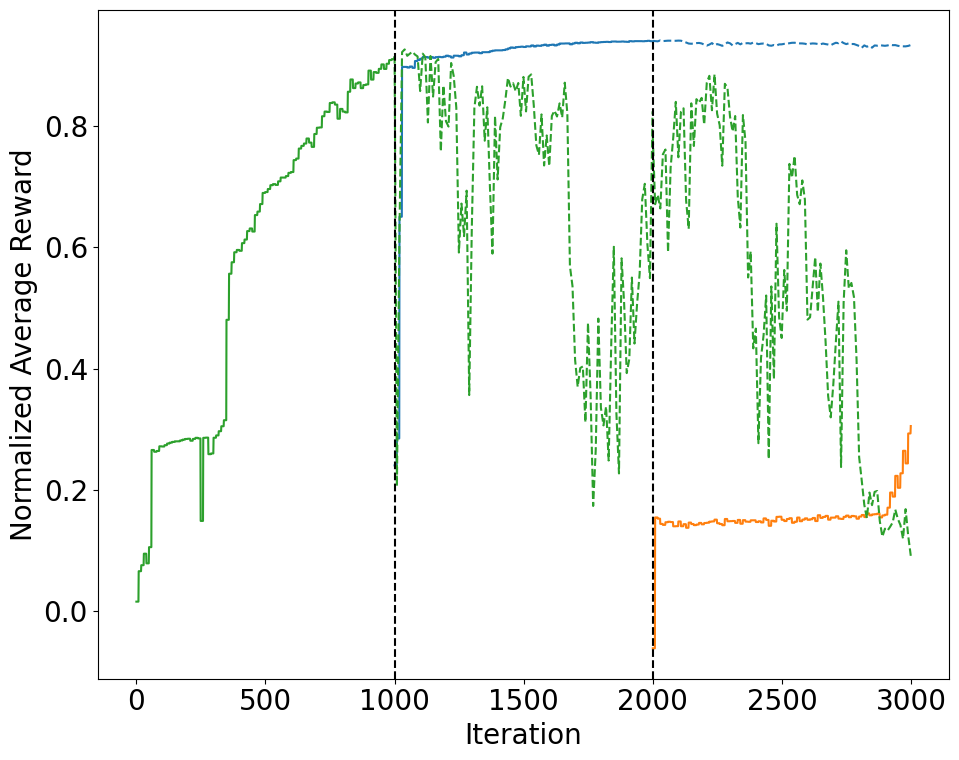}
    \includegraphics[width=\linewidth]{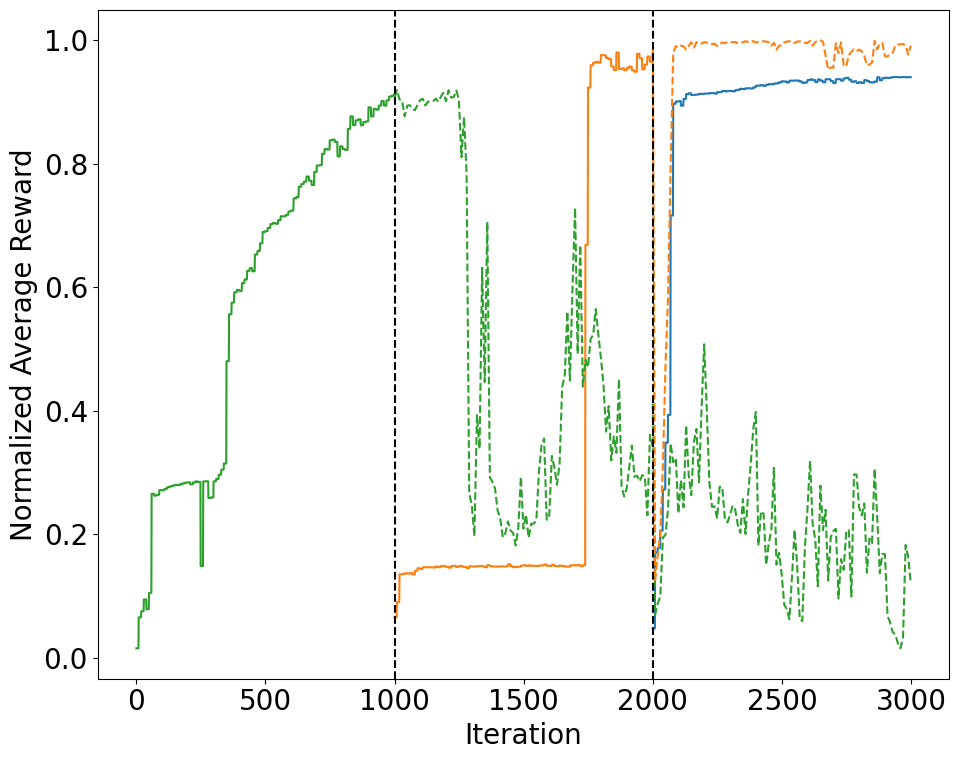}
\end{subfigure}

\caption{Sequential continual learning experiments for the three MuJoCo environments. Each plot corresponds to a different task ordering, showing the normalized average reward over training iterations when the policy is trained sequentially on all tasks using a shared output head (\textit{Hopper-v5}, \textit{Swimmer-v5}, and \textit{Walker2d-v5}). Vertical dashed lines mark task transitions. Normalization is done with respect to the maximum reward achieved in the corresponding single-task run.}
\Description{Sequential continual learning experiments for the three MuJoCo environments. Each plot corresponds to a different task ordering, showing the normalized average reward over training iterations when the policy is trained sequentially on all tasks using a shared output head (\textit{Hopper-v5}, \textit{Swimmer-v5}, and \textit{Walker2d-v5}). Vertical dashed lines mark task transitions. Normalization is done with respect to the maximum reward achieved in the corresponding single-task run.}
\label{fig:sequential_shared}
\end{figure*}

\begin{figure*}[p]
\centering
\setkeys{Gin}{height=0.23\textheight,keepaspectratio}

\begin{subfigure}{0.45\textwidth}
    \centering
    \includegraphics[width=\linewidth]{distinct_hsw_first.png}
    \includegraphics[width=\linewidth]{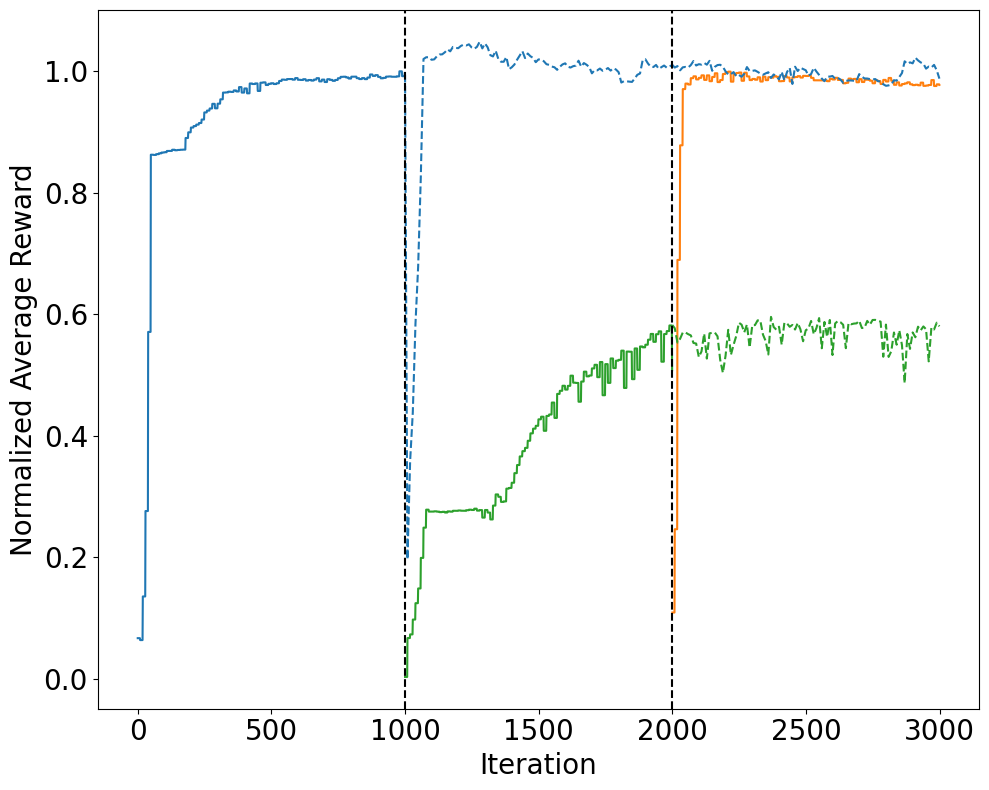}
    \includegraphics[width=\linewidth]{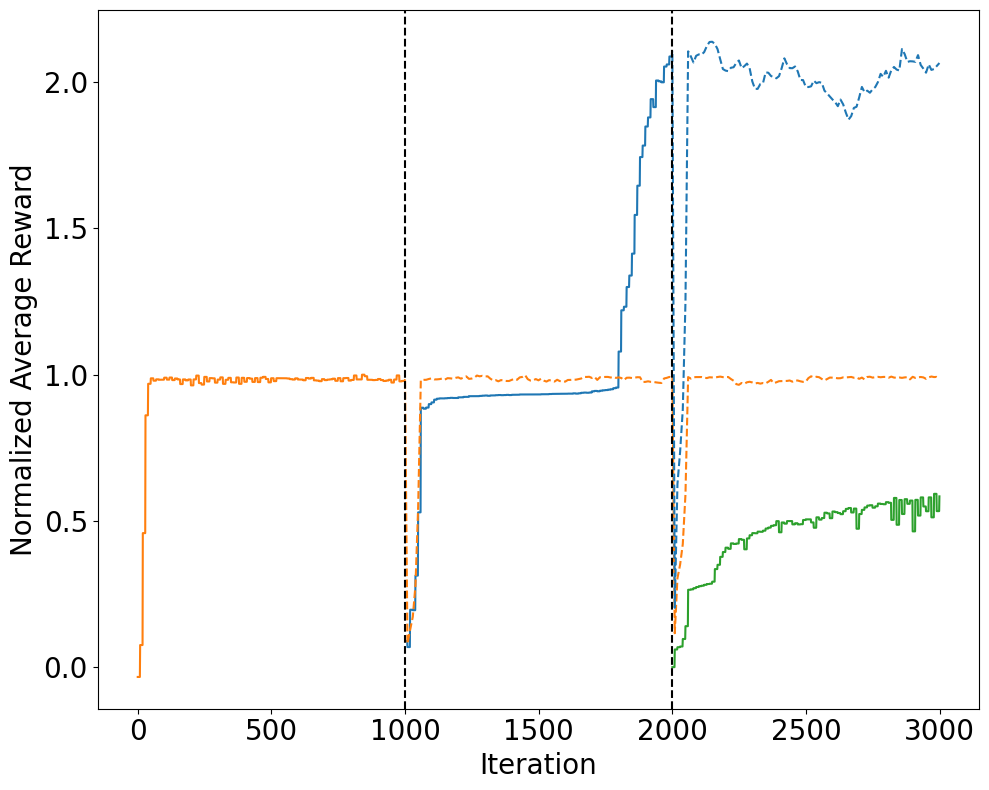}
\end{subfigure}\hfill
\begin{subfigure}{0.45\textwidth}
    \centering
    \includegraphics[width=\linewidth]{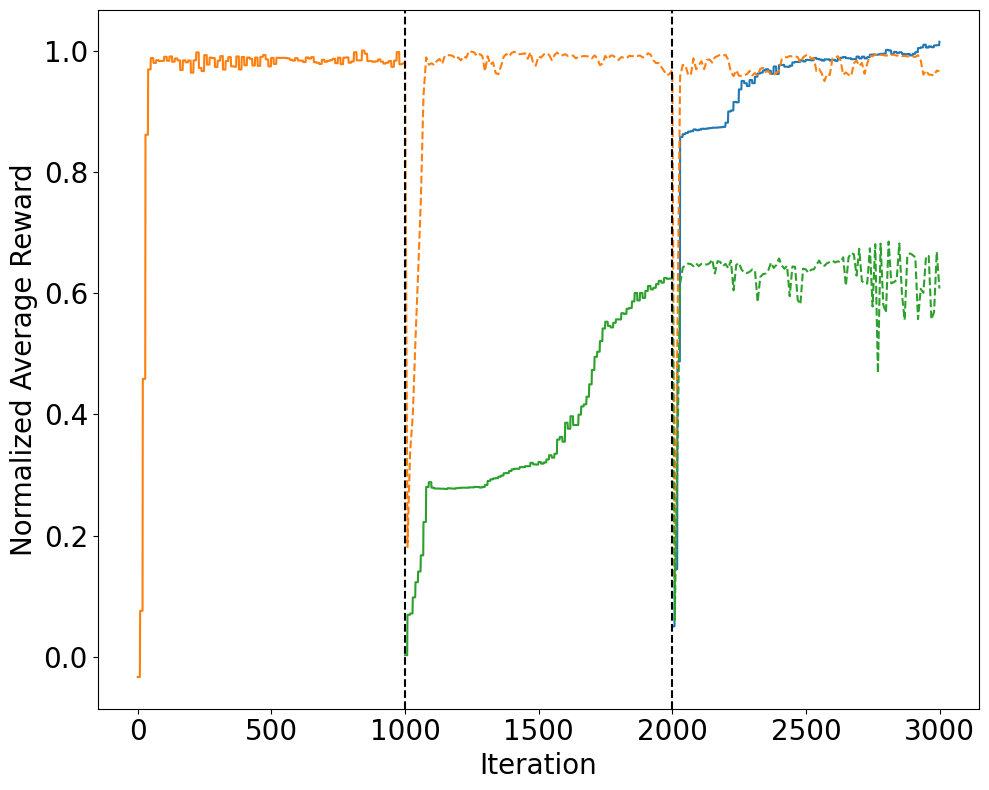}
    \includegraphics[width=\linewidth]{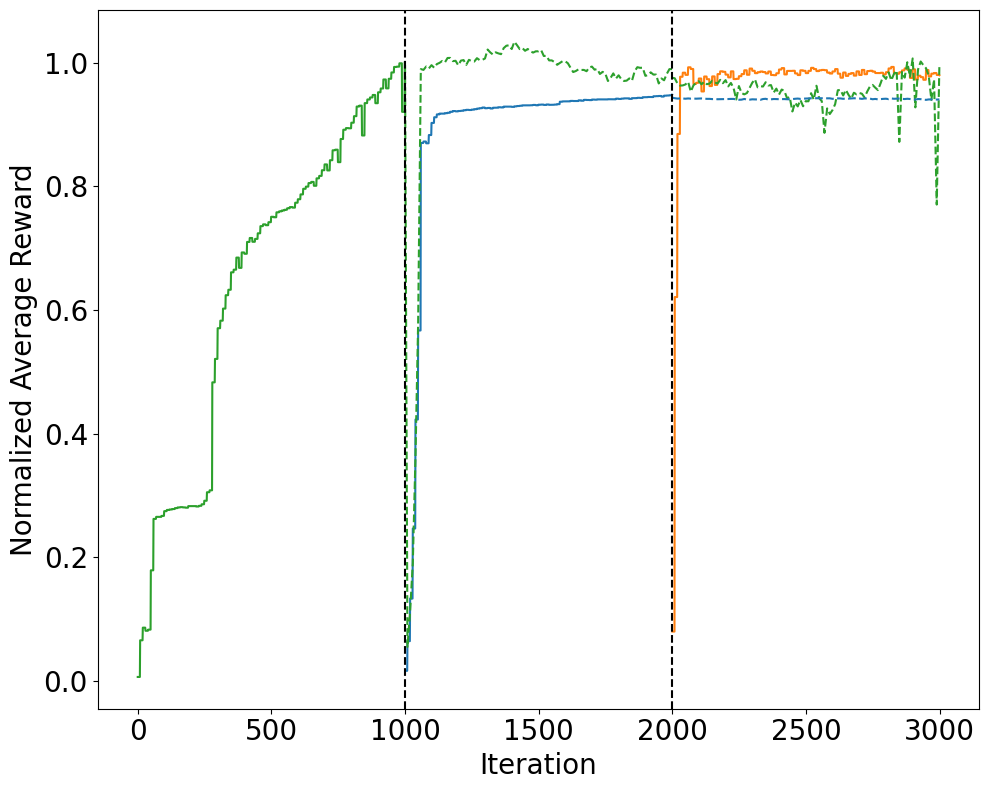}
    \includegraphics[width=\linewidth]{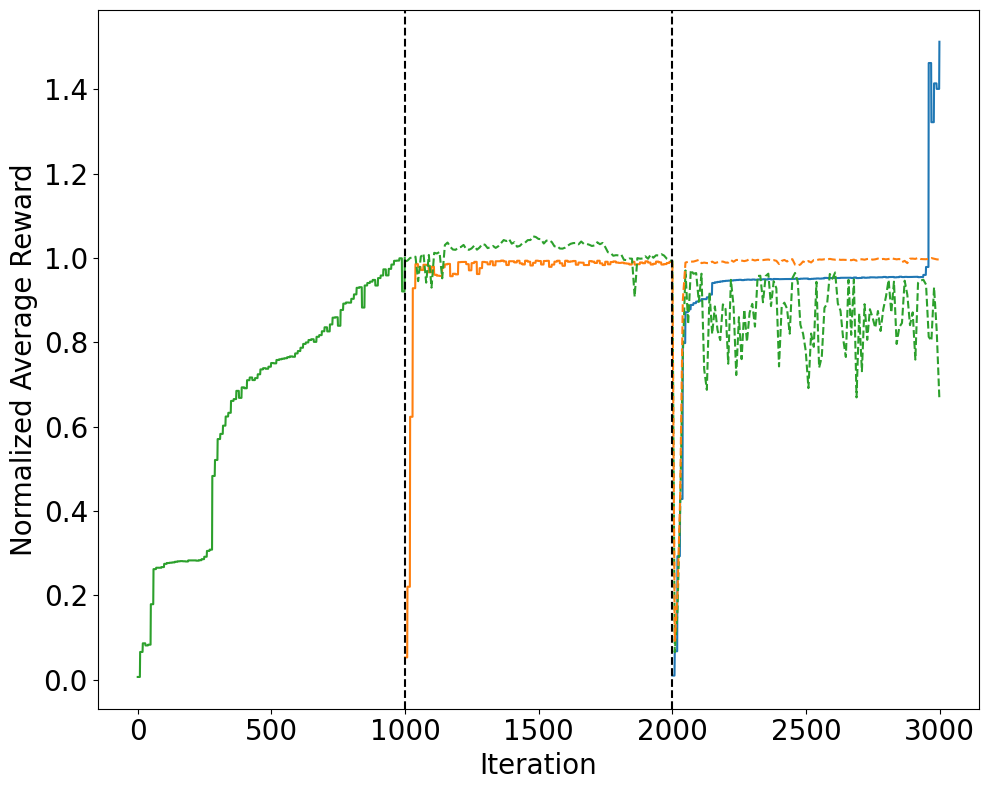}
\end{subfigure}

\caption{Sequential continual learning experiments for the three MuJoCo environments. Each plot corresponds to a different task ordering, showing the normalized average reward over training iterations when the policy is trained sequentially on all tasks using a distinct output head (\textit{Hopper-v5}, \textit{Swimmer-v5}, and \textit{Walker2d-v5}). Vertical dashed lines mark task transitions. Normalization is done with respect to the maximum reward achieved in the corresponding single-task run.}
\Description{Sequential continual learning experiments for the three MuJoCo environments. Each plot corresponds to a different task ordering, showing the normalized average reward over training iterations when the policy is trained sequentially on all tasks using a distinct output head (\textit{Hopper-v5}, \textit{Swimmer-v5}, and \textit{Walker2d-v5}). Vertical dashed lines mark task transitions. Normalization is done with respect to the maximum reward achieved in the corresponding single-task run.}
\label{fig:sequential_distinct}
\end{figure*}

\begin{figure*}[p]
\centering
\setkeys{Gin}{height=0.23\textheight,keepaspectratio}

\begin{subfigure}{0.45\textwidth}
    \centering
    \includegraphics[width=\linewidth]{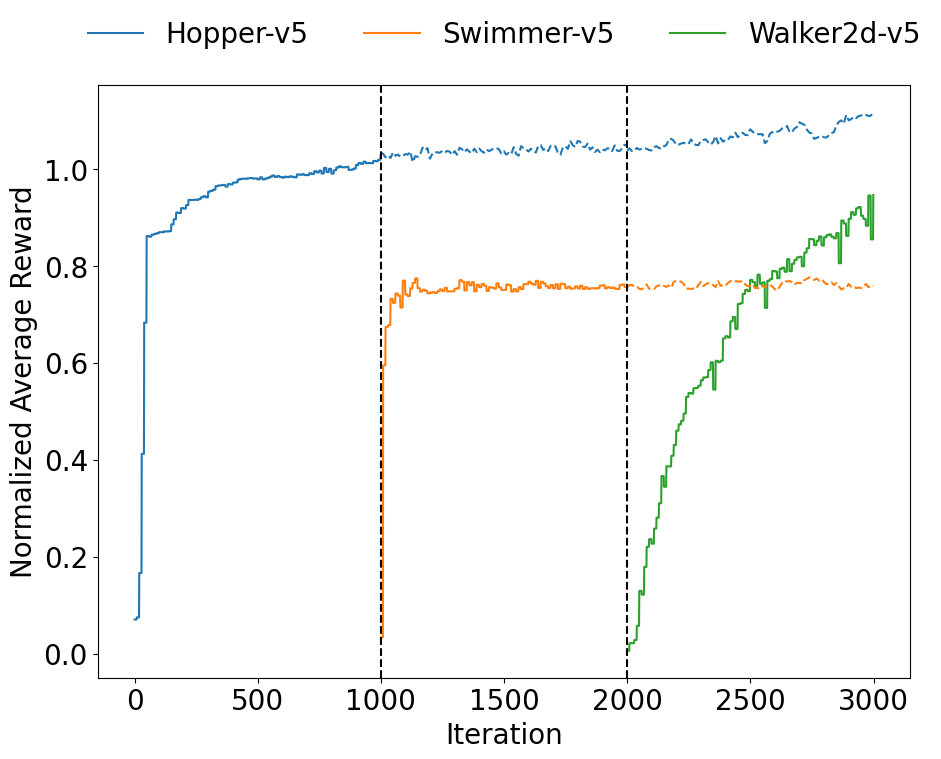}
    \includegraphics[width=\linewidth]{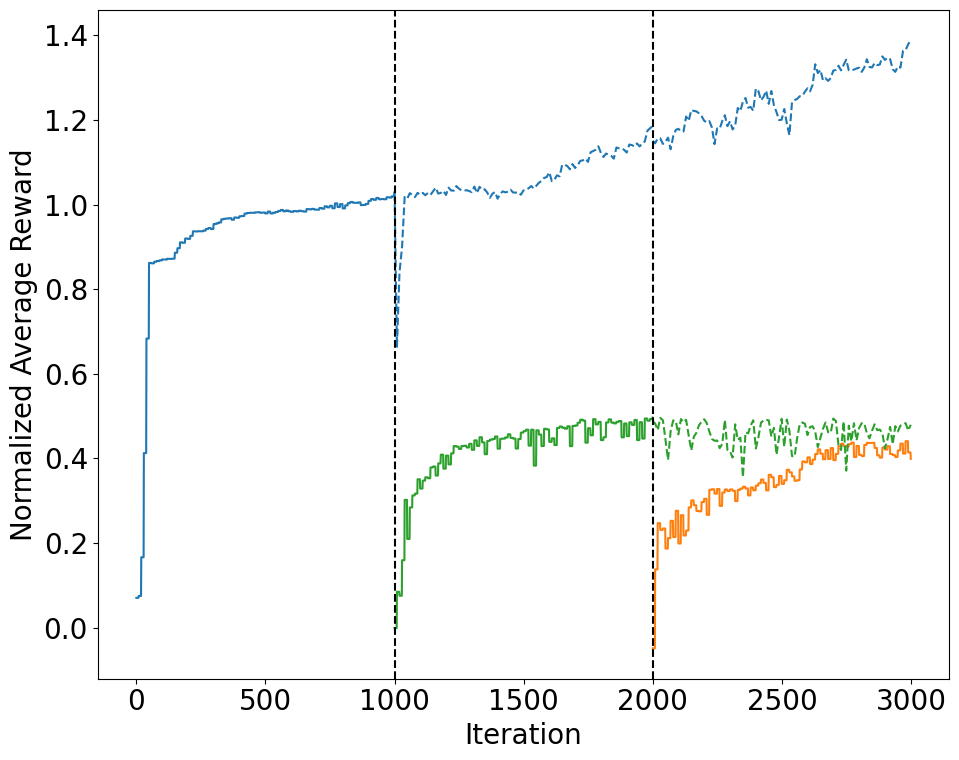}
    \includegraphics[width=\linewidth]{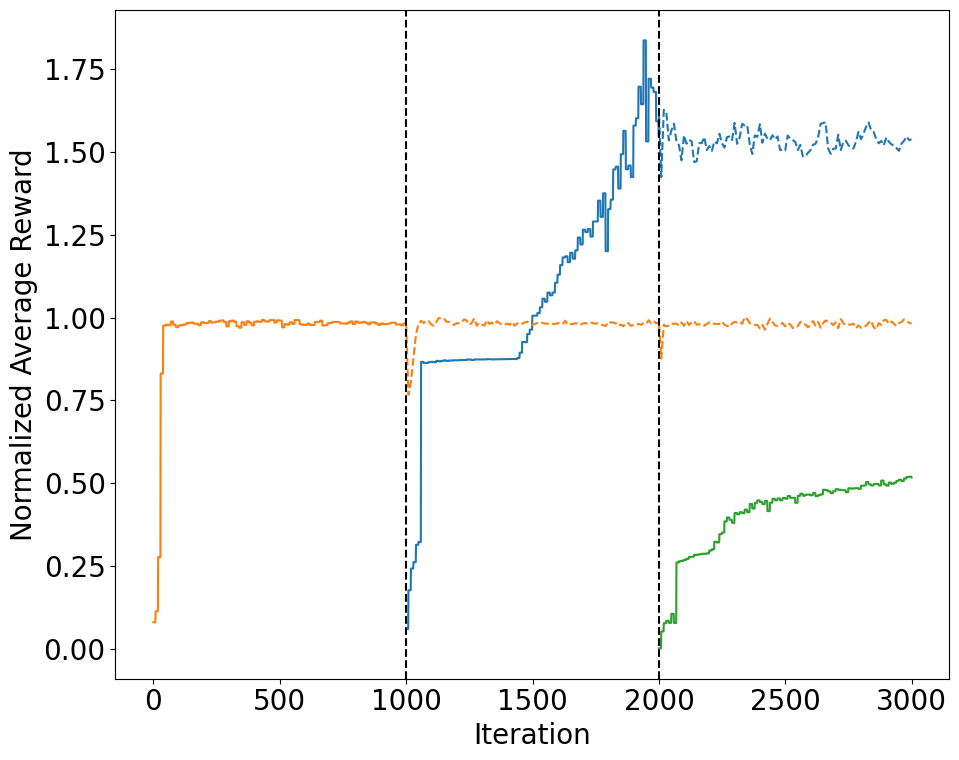}
\end{subfigure}\hfill
\begin{subfigure}{0.45\textwidth}
    \centering
    \includegraphics[width=\linewidth]{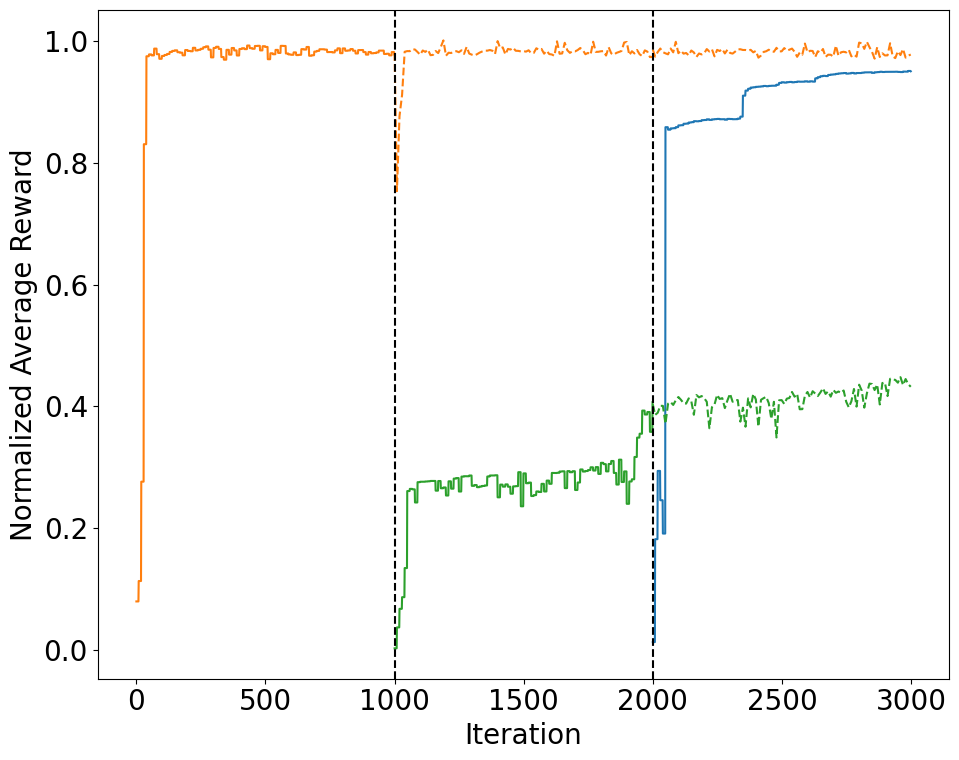}
    \includegraphics[width=\linewidth]{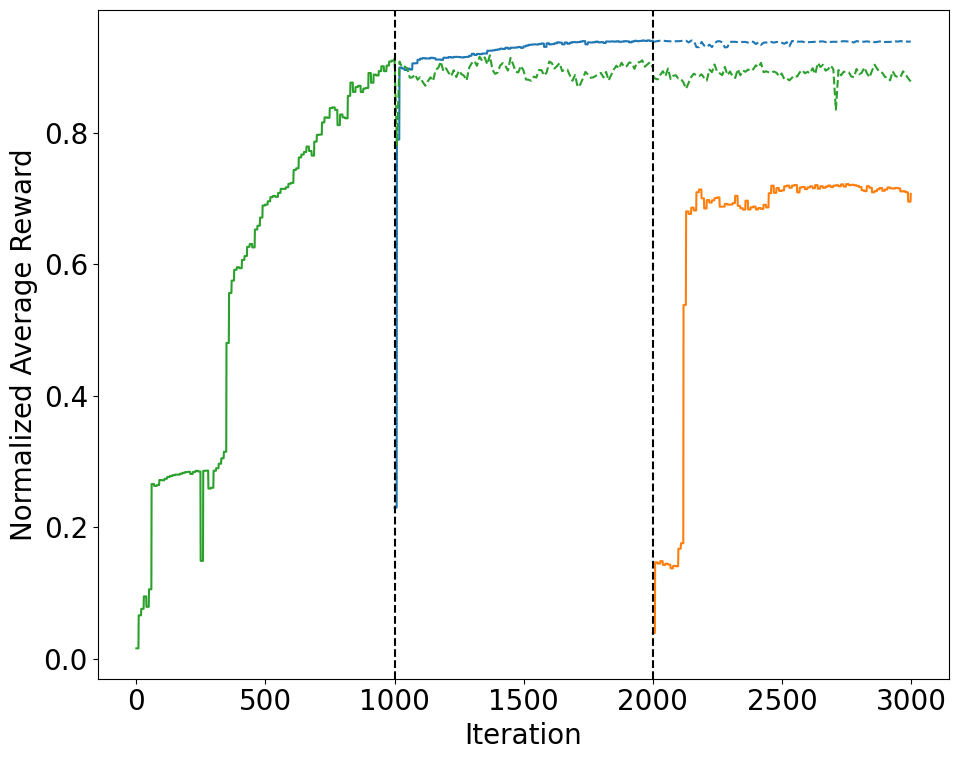}
    \includegraphics[width=\linewidth]{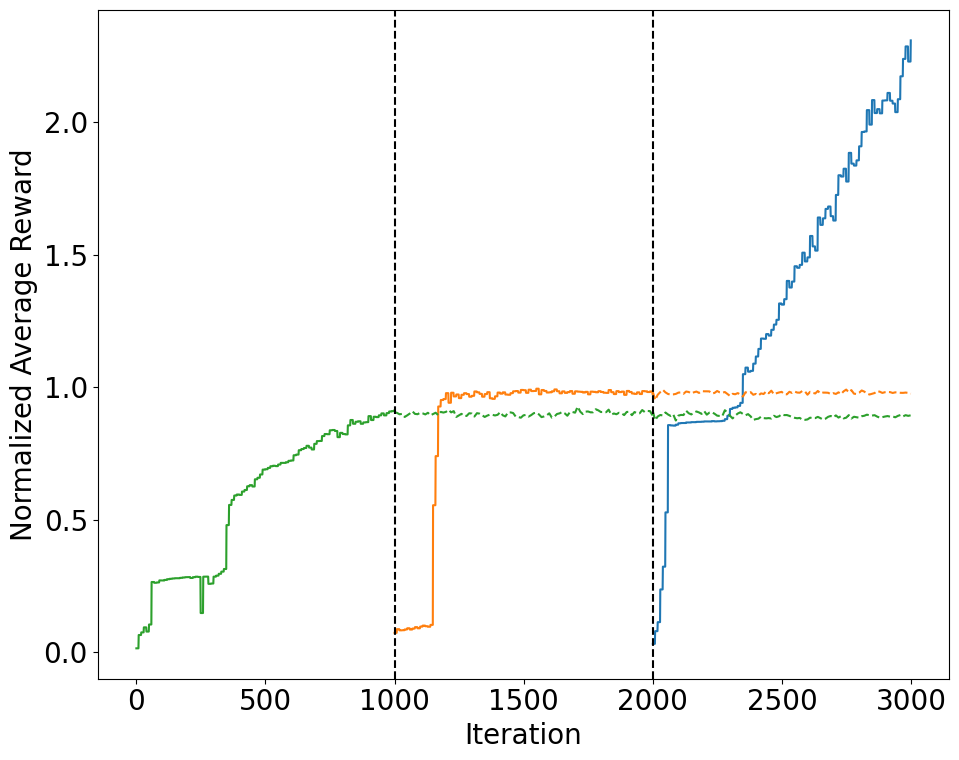}
\end{subfigure}

\caption{Sequential continual learning experiments with the replay mechanism. Each plot shows the normalized average reward over training iterations for one task ordering, where the policy is trained sequentially on all tasks using a shared output head (\textit{Hopper-v5}, \textit{Swimmer-v5}, and \textit{Walker2d-v5}). Replay is performed with 192 added perturbations (approximately 25\% of the population size). Vertical dashed lines mark task transitions. Normalization is done with respect to the maximum reward achieved in the corresponding single-task run.}
\Description{Sequential continual learning experiments with the replay mechanism. Each plot shows the normalized average reward over training iterations for one task ordering, where the policy is trained sequentially on all tasks using a shared output head (\textit{Hopper-v5}, \textit{Swimmer-v5}, and \textit{Walker2d-v5}). Replay is performed with 192 added perturbations (approximately 25\% of the population size). Vertical dashed lines mark task transitions. Normalization is done with respect to the maximum reward achieved in the corresponding single-task run.}
\label{fig:sequential_replay}
\end{figure*}

\begin{figure*}[p]
\centering
\setkeys{Gin}{height=0.23\textheight,keepaspectratio}

\begin{subfigure}{0.45\textwidth}
    \centering
    \includegraphics[width=\linewidth]{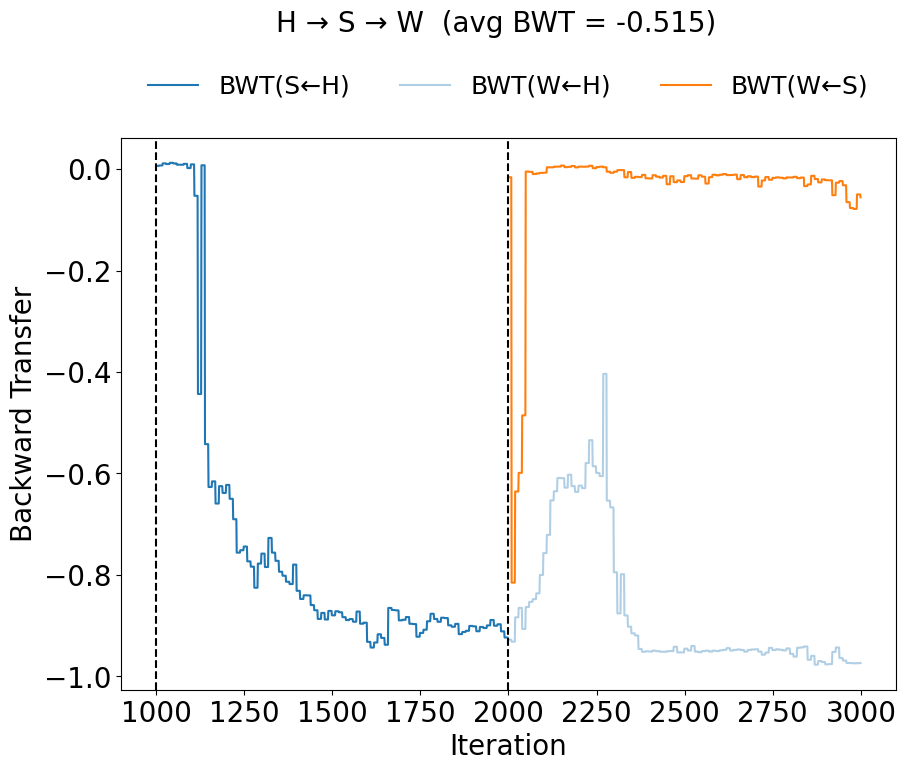}
    \includegraphics[width=\linewidth]{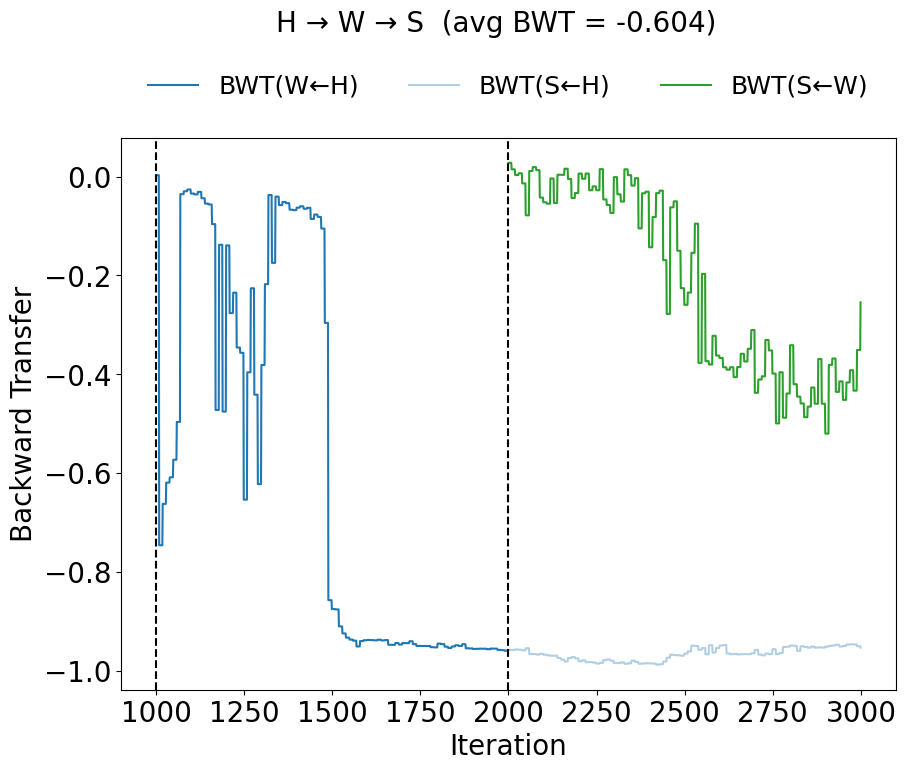}
    \includegraphics[width=\linewidth]{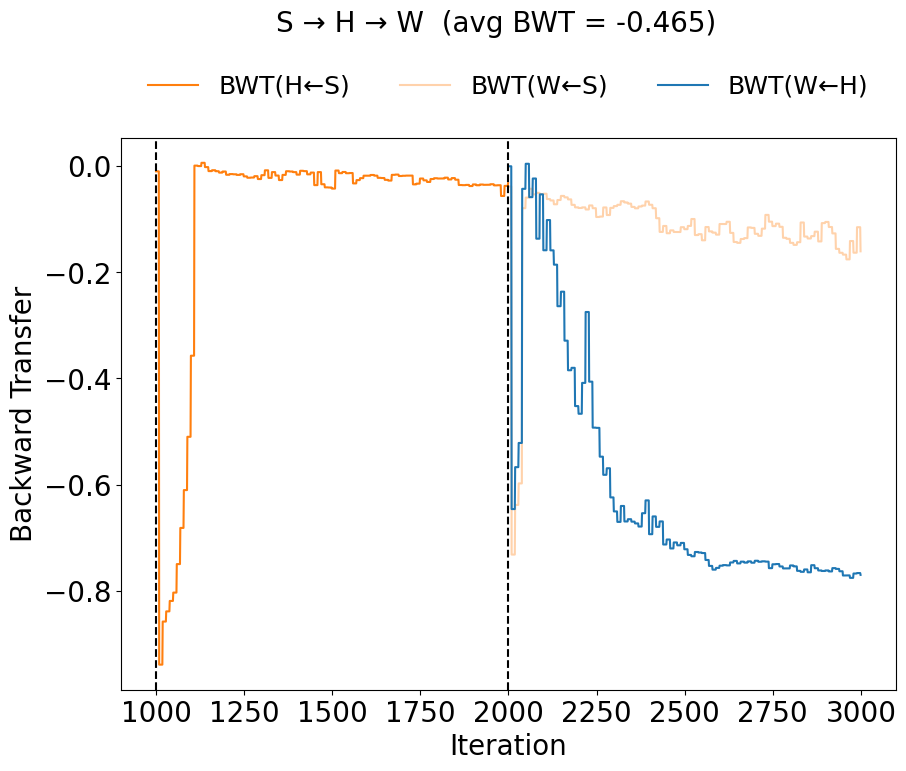}
\end{subfigure}\hfill
\begin{subfigure}{0.45\textwidth}
    \centering
    \includegraphics[width=\linewidth]{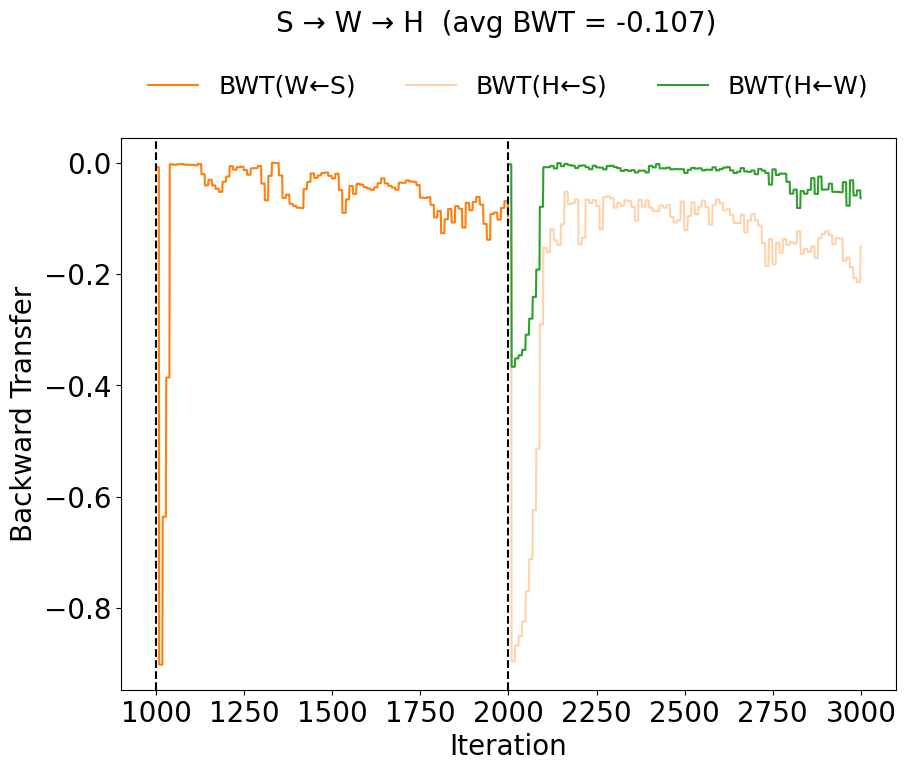}
    \includegraphics[width=\linewidth]{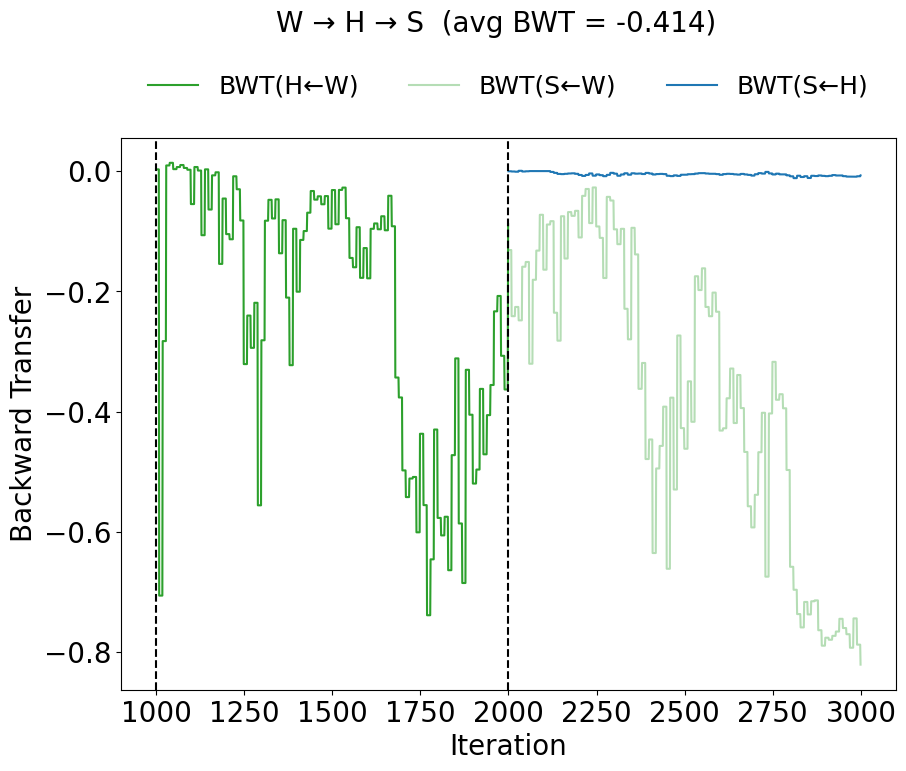}
    \includegraphics[width=\linewidth]{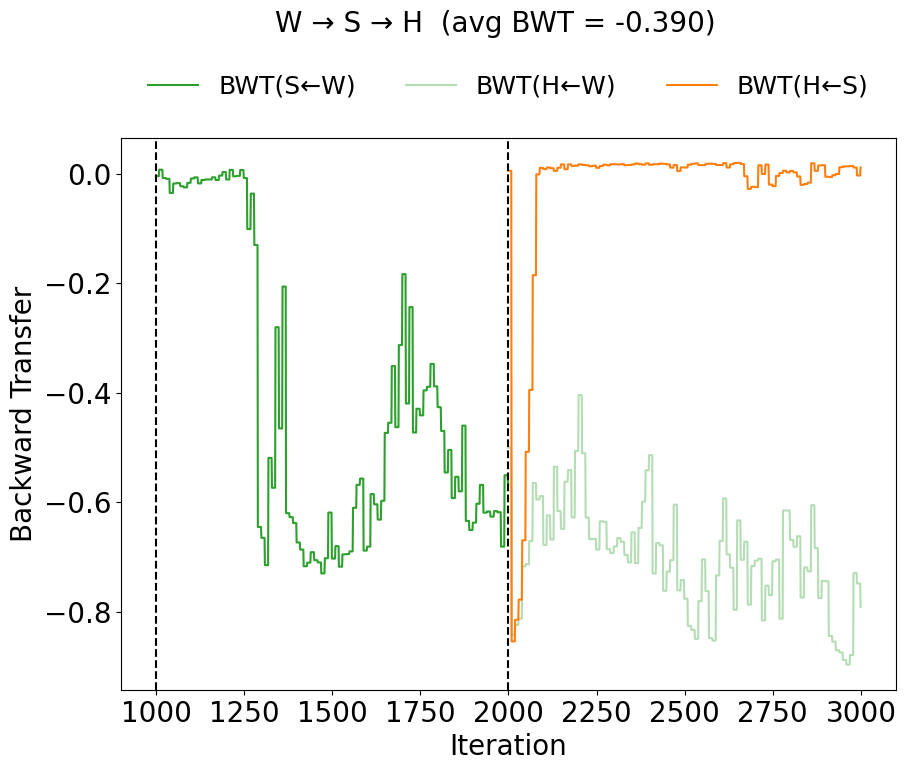}
\end{subfigure}

\caption{Backward transfer during sequential continual learning for the shared output head policies. Each plot corresponds to a different task ordering across the three MuJoCo environments (\textit{Hopper-v5}, \textit{Swimmer-v5}, and \textit{Walker2d-v5}). Curves show the evolution of task-wise backward transfer (BWT) values over training iterations, with vertical dashed lines marking task transitions. Normalization is done with respect to the maximum reward achieved in the corresponding single-task runs.}
\Description{Backward transfer during sequential continual learning for the shared output head policies. Each plot corresponds to a different task ordering across the three MuJoCo environments (\textit{Hopper-v5}, \textit{Swimmer-v5}, and \textit{Walker2d-v5}). Curves show the evolution of task-wise backward transfer (BWT) values over training iterations, with vertical dashed lines marking task transitions. Normalization is done with respect to the maximum reward achieved in the corresponding single-task runs.}
\label{fig:bwt_shared}
\end{figure*}

\begin{figure*}[p]
\centering
\setkeys{Gin}{height=0.23\textheight,keepaspectratio}

\begin{subfigure}{0.45\textwidth}
    \centering
    \includegraphics[width=\linewidth]{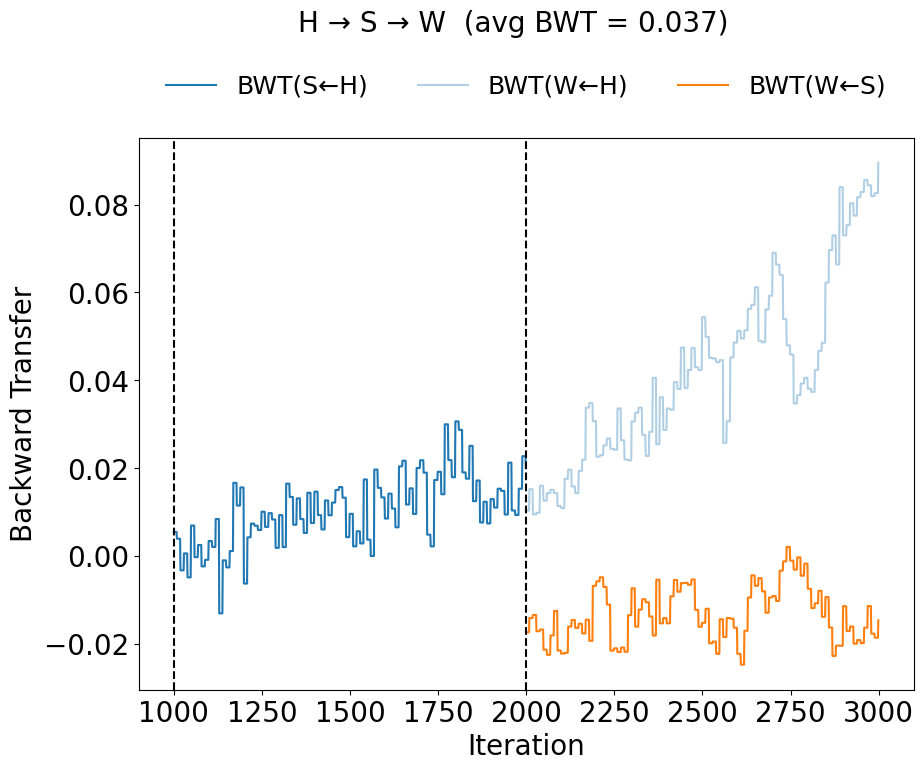}
    \includegraphics[width=\linewidth]{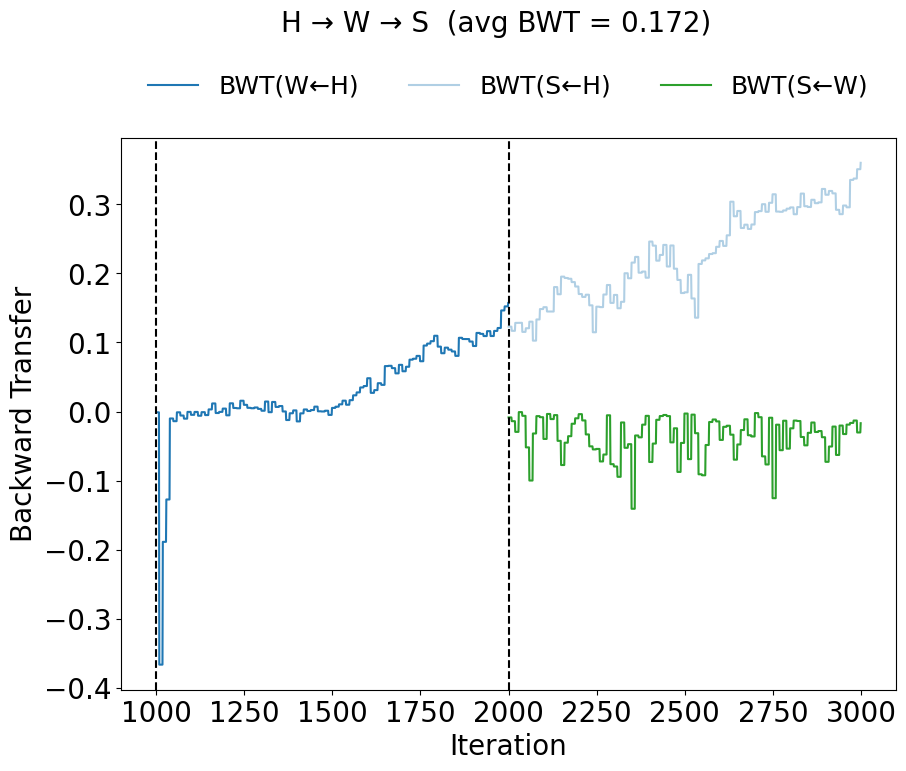}
    \includegraphics[width=\linewidth]{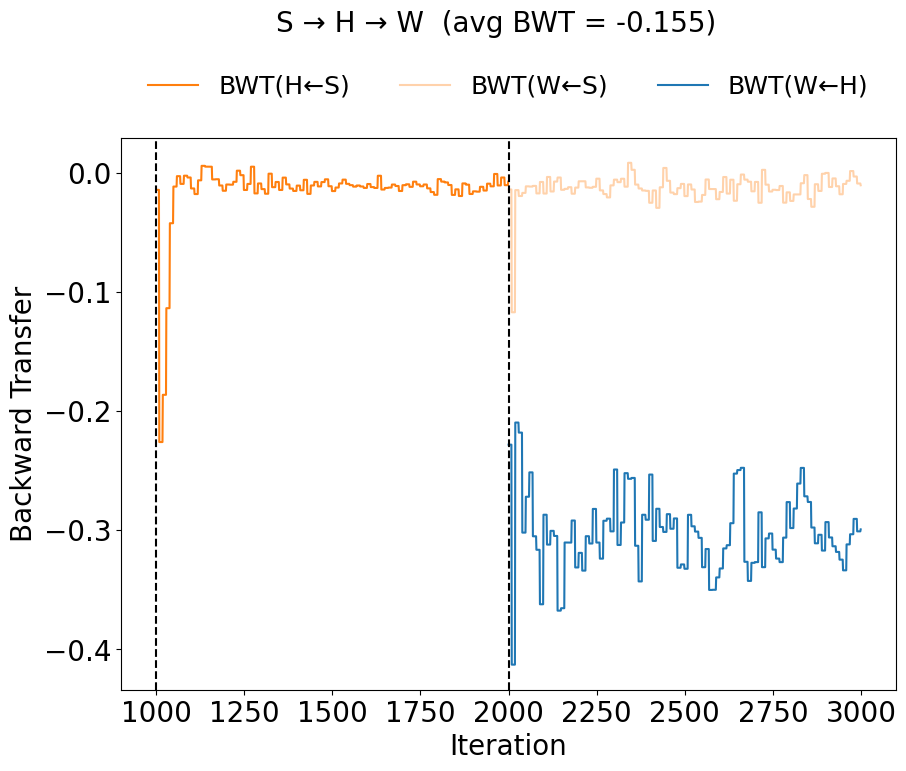}
\end{subfigure}\hfill
\begin{subfigure}{0.45\textwidth}
    \centering
    \includegraphics[width=\linewidth]{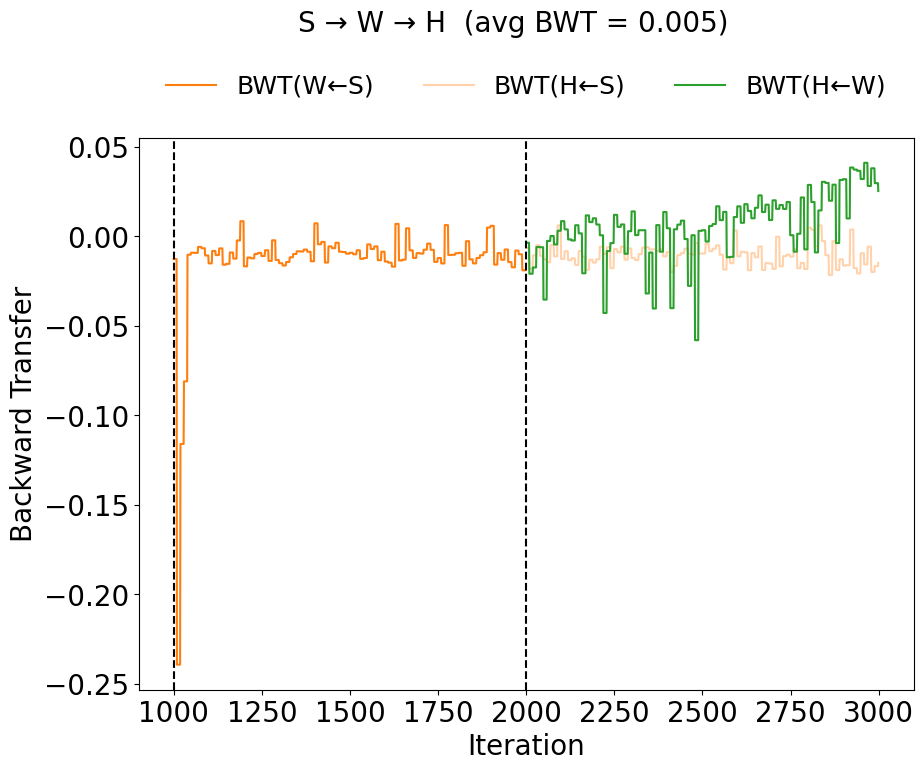}
    \includegraphics[width=\linewidth]{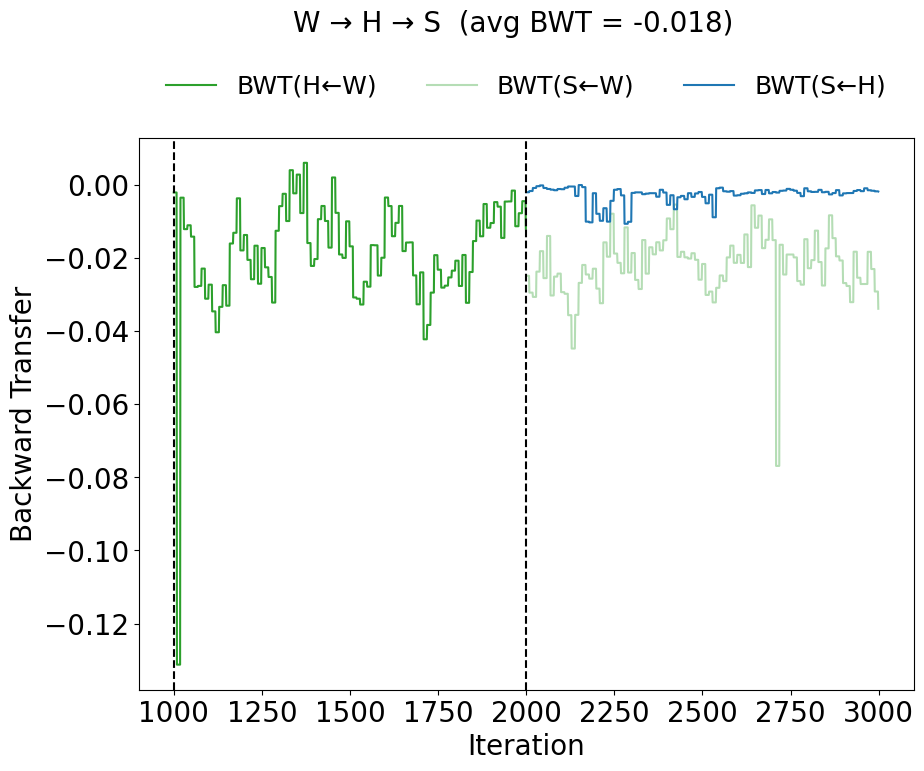}
    \includegraphics[width=\linewidth]{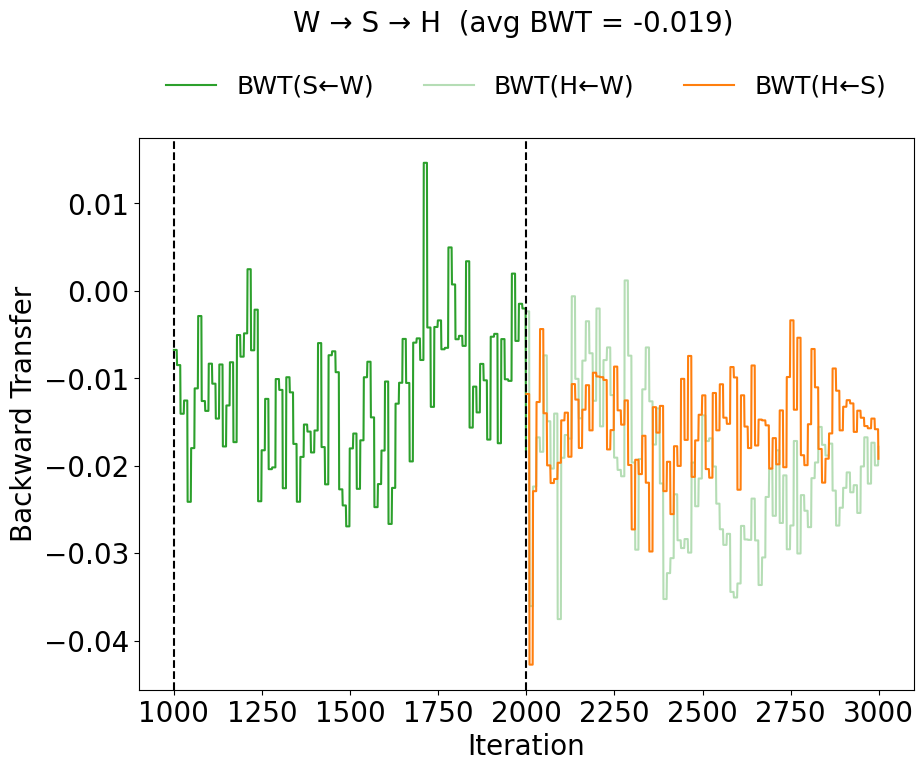}
\end{subfigure}

\caption{Backward transfer during sequential continual learning for the shared output head policies, supported with replay steps. Each plot corresponds to a different task ordering across the three MuJoCo environments (\textit{Hopper-v5}, \textit{Swimmer-v5}, and \textit{Walker2d-v5}). Curves show the evolution of task-wise backward transfer (BWT) values over training iterations, with vertical dashed lines marking task transitions. Normalization is done with respect to the maximum reward achieved in the corresponding single-task runs.}
\Description{Backward transfer during sequential continual learning for the shared output head policies, supported with replay steps. Each plot corresponds to a different task ordering across the three MuJoCo environments (\textit{Hopper-v5}, \textit{Swimmer-v5}, and \textit{Walker2d-v5}). Curves show the evolution of task-wise backward transfer (BWT) values over training iterations, with vertical dashed lines marking task transitions. Normalization is done with respect to the maximum reward achieved in the corresponding single-task runs.}
\label{fig:bwt_replay}
\end{figure*}

\begin{figure*}[p]
\centering
\setkeys{Gin}{height=0.23\textheight,keepaspectratio}

\begin{subfigure}{0.45\textwidth}
    \centering
    \includegraphics[width=\linewidth]{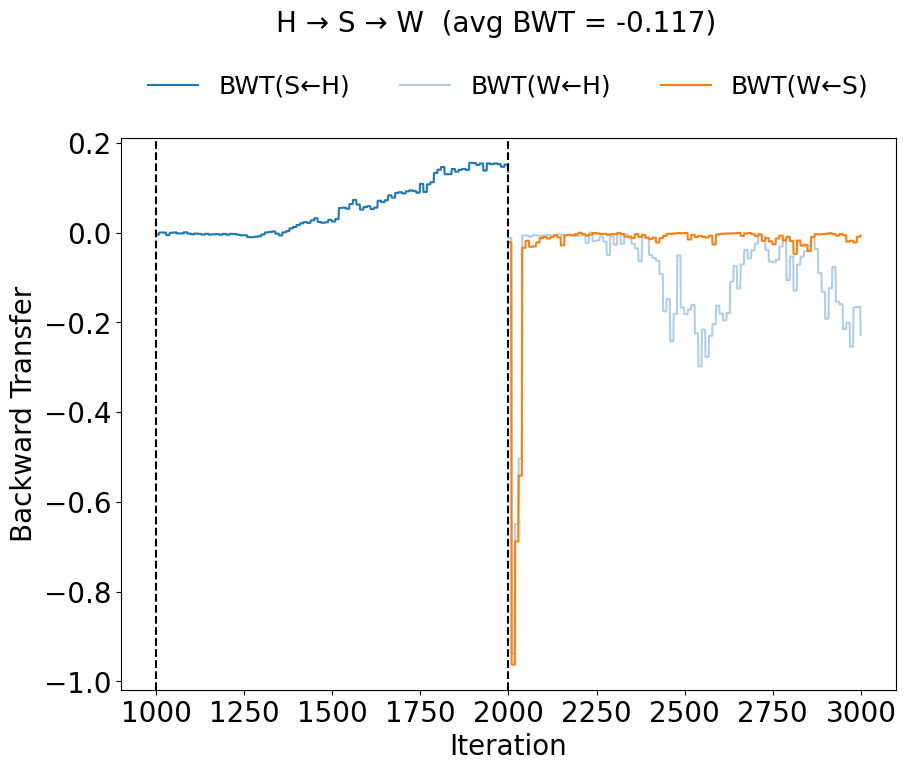}
    \includegraphics[width=\linewidth]{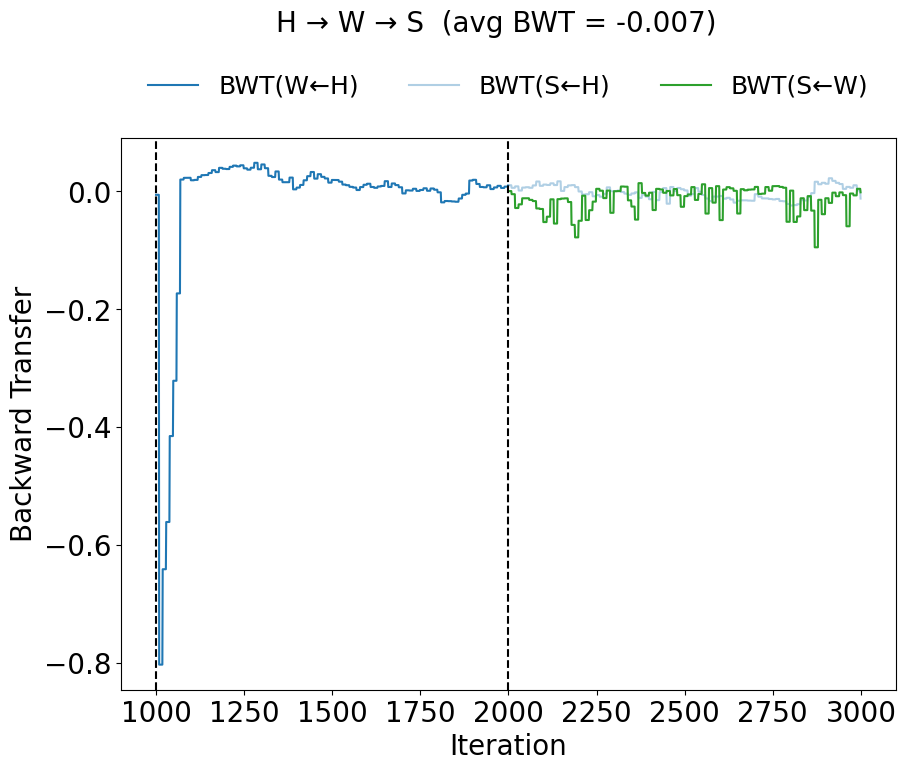}
    \includegraphics[width=\linewidth]{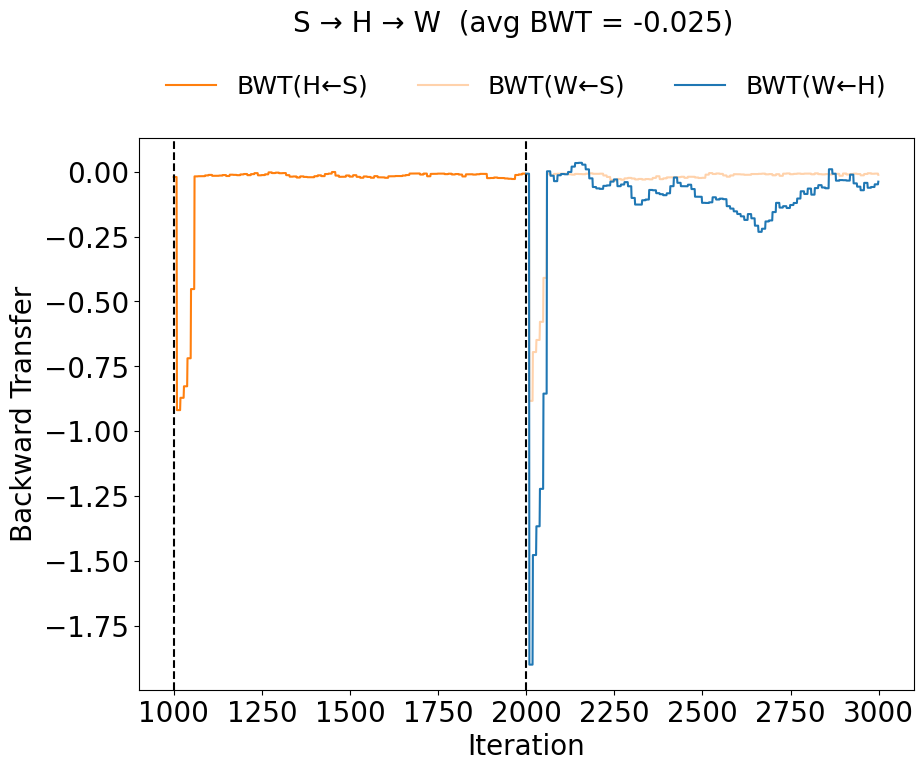}
\end{subfigure}\hfill
\begin{subfigure}{0.45\textwidth}
    \centering
    \includegraphics[width=\linewidth]{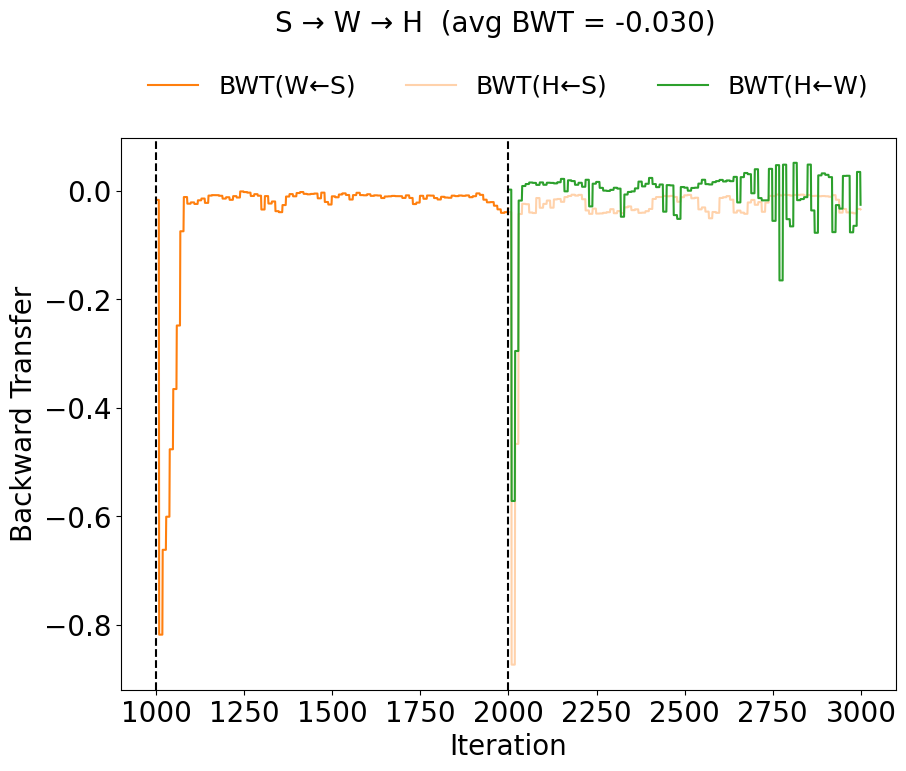}
    \includegraphics[width=\linewidth]{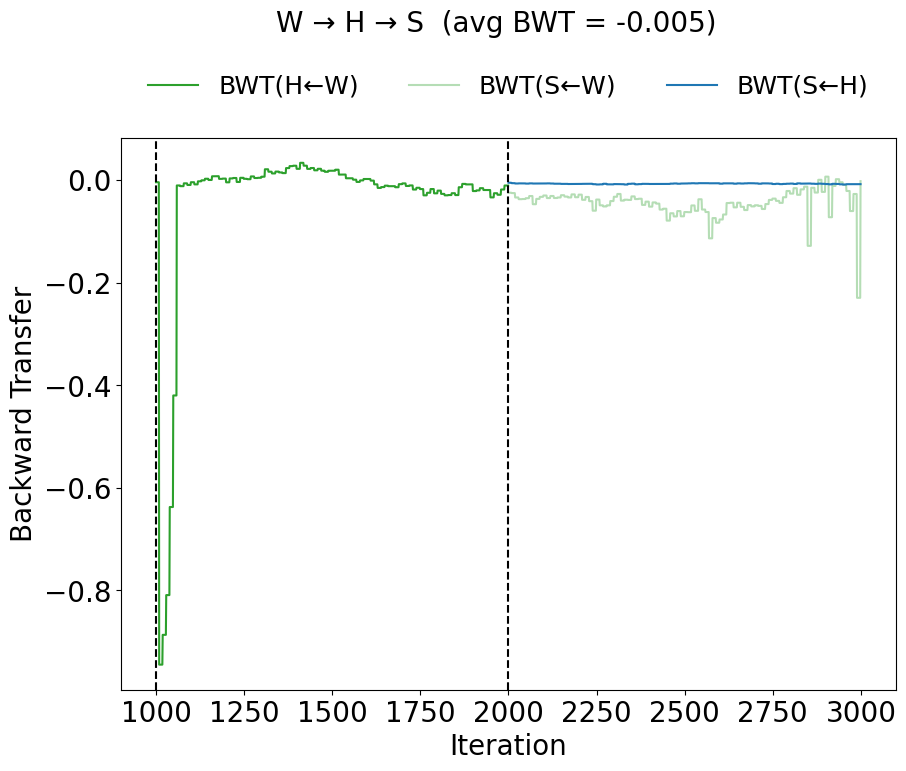}
    \includegraphics[width=\linewidth]{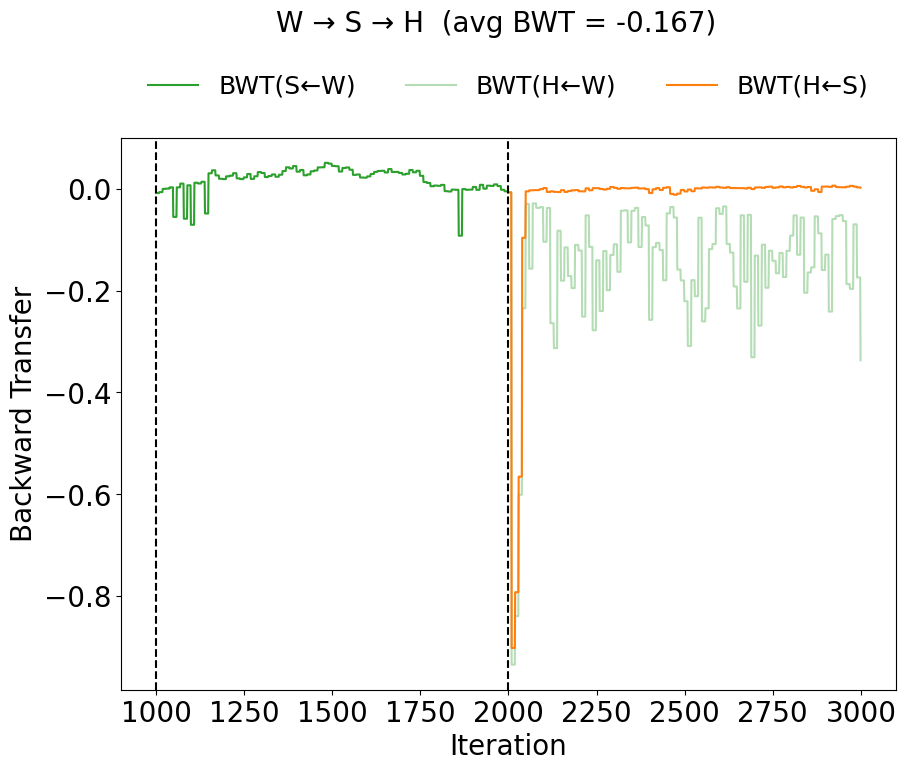}
\end{subfigure}

\caption{Backward transfer during sequential continual learning for the distinct output head policies. Each plot corresponds to a different task ordering across the three MuJoCo environments (\textit{Hopper-v5}, \textit{Swimmer-v5}, and \textit{Walker2d-v5}). Curves show the evolution of task-wise backward transfer (BWT) values over training iterations, with vertical dashed lines marking task transitions. Normalization is done with respect to the maximum reward achieved in the corresponding single-task runs.}
\Description{Backward transfer during sequential continual learning for the distinct output head policies. Each plot corresponds to a different task ordering across the three MuJoCo environments (\textit{Hopper-v5}, \textit{Swimmer-v5}, and \textit{Walker2d-v5}). Curves show the evolution of task-wise backward transfer (BWT) values over training iterations, with vertical dashed lines marking task transitions. Normalization is done with respect to the maximum reward achieved in the corresponding single-task runs.}
\label{fig:bwt_distinct}
\end{figure*}

\begin{figure*}[p]
\centering
\setkeys{Gin}{height=0.23\textheight,keepaspectratio}

\begin{subfigure}{0.45\textwidth}
    \centering
    \includegraphics[width=\linewidth]{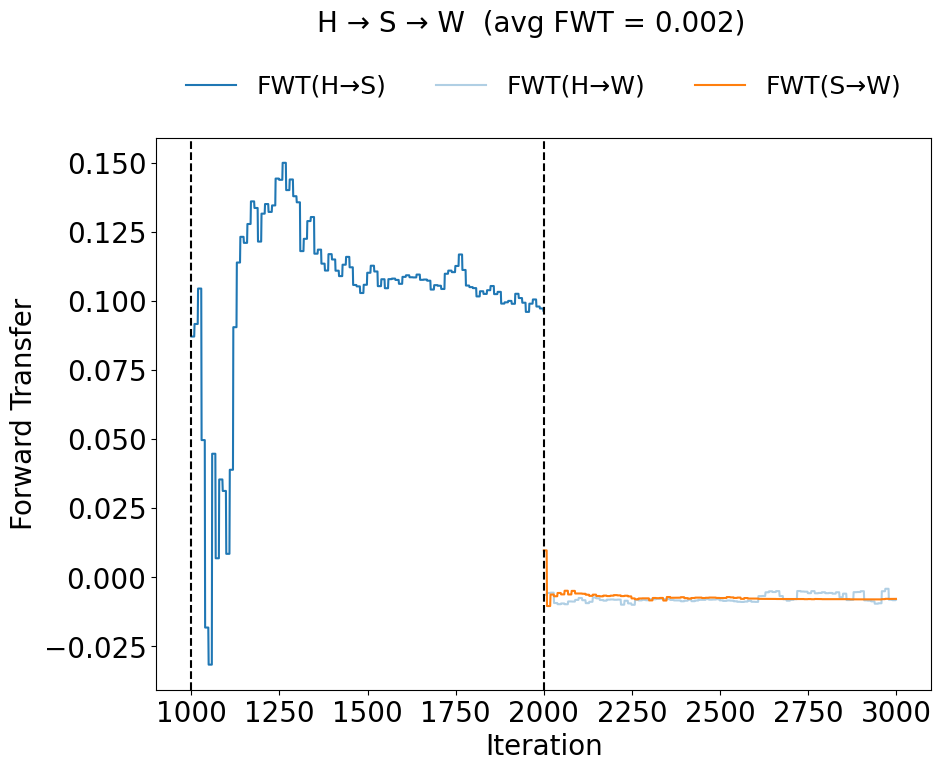}
    \includegraphics[width=\linewidth]{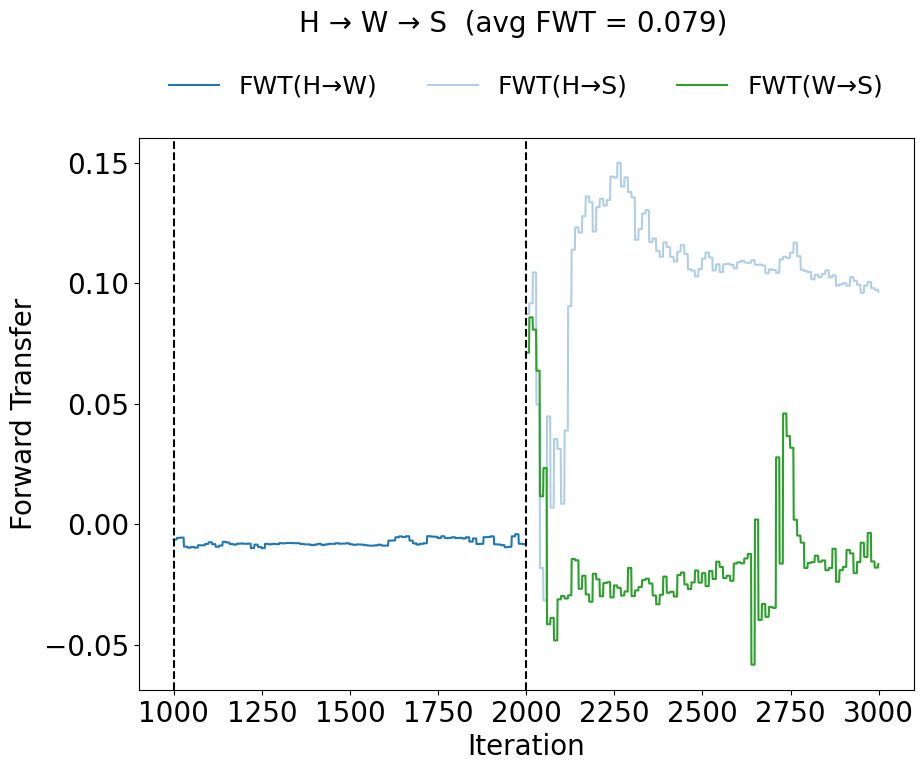}
    \includegraphics[width=\linewidth]{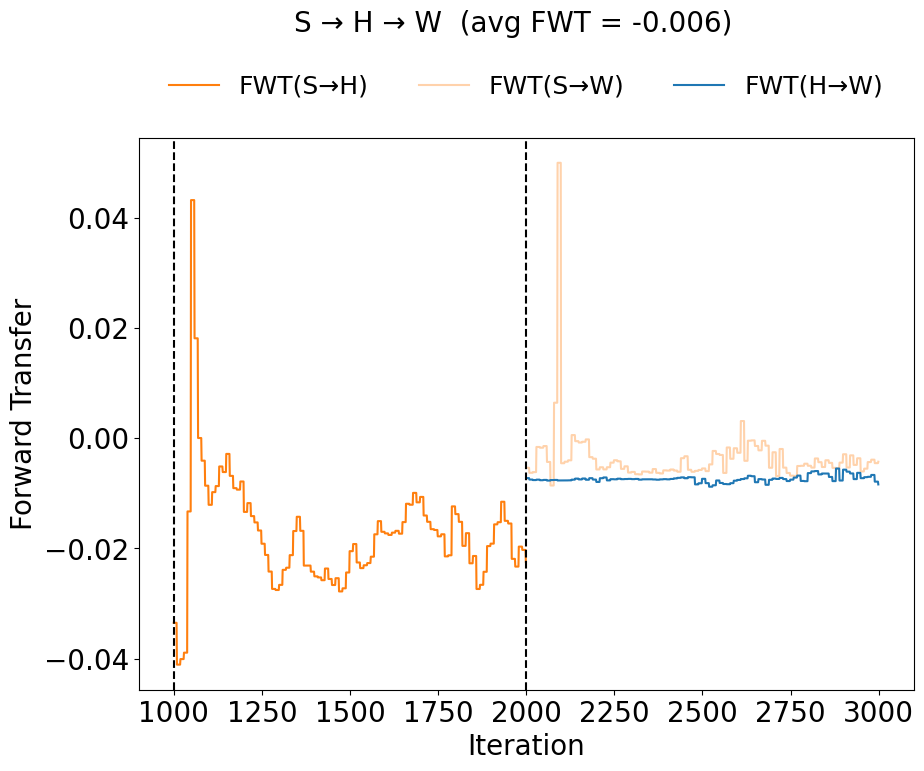}
\end{subfigure}\hfill
\begin{subfigure}{0.45\textwidth}
    \centering
    \includegraphics[width=\linewidth]{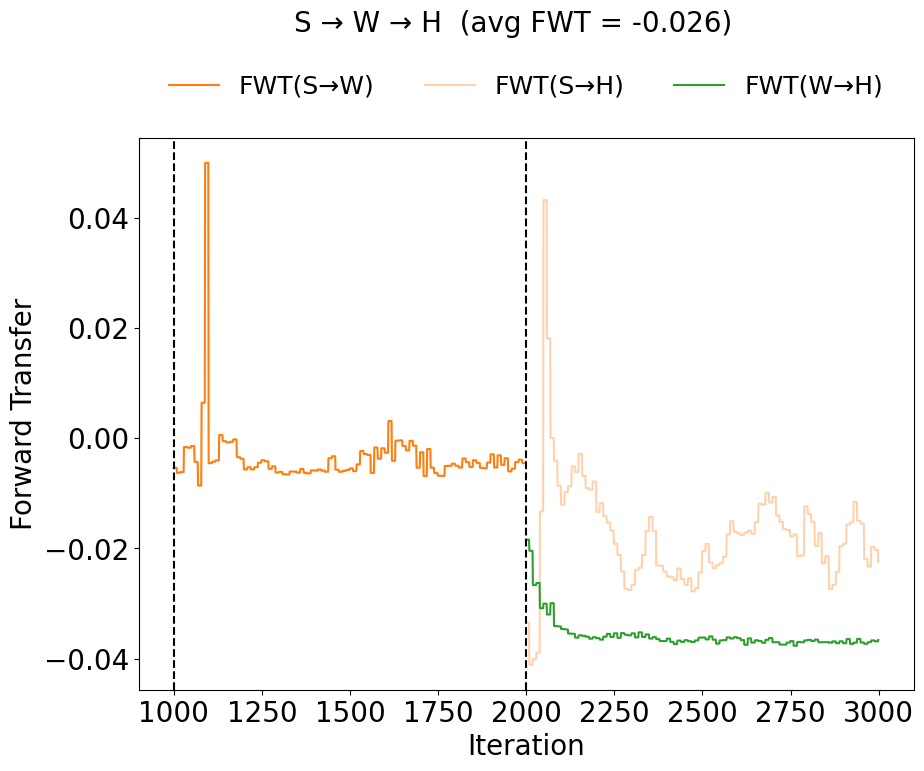}
    \includegraphics[width=\linewidth]{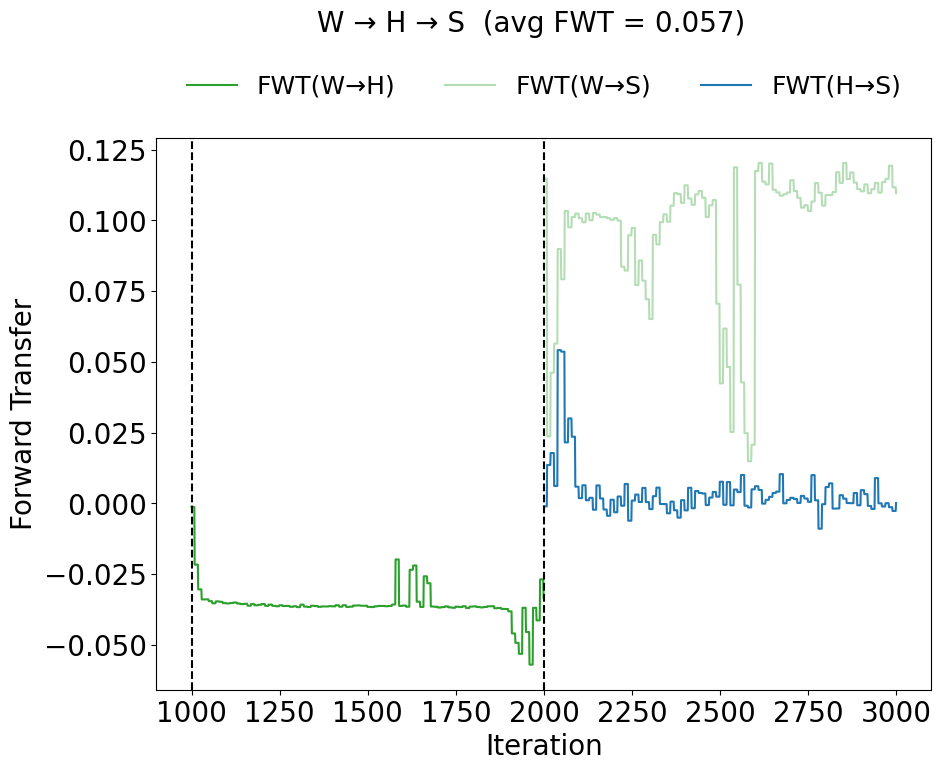}
    \includegraphics[width=\linewidth]{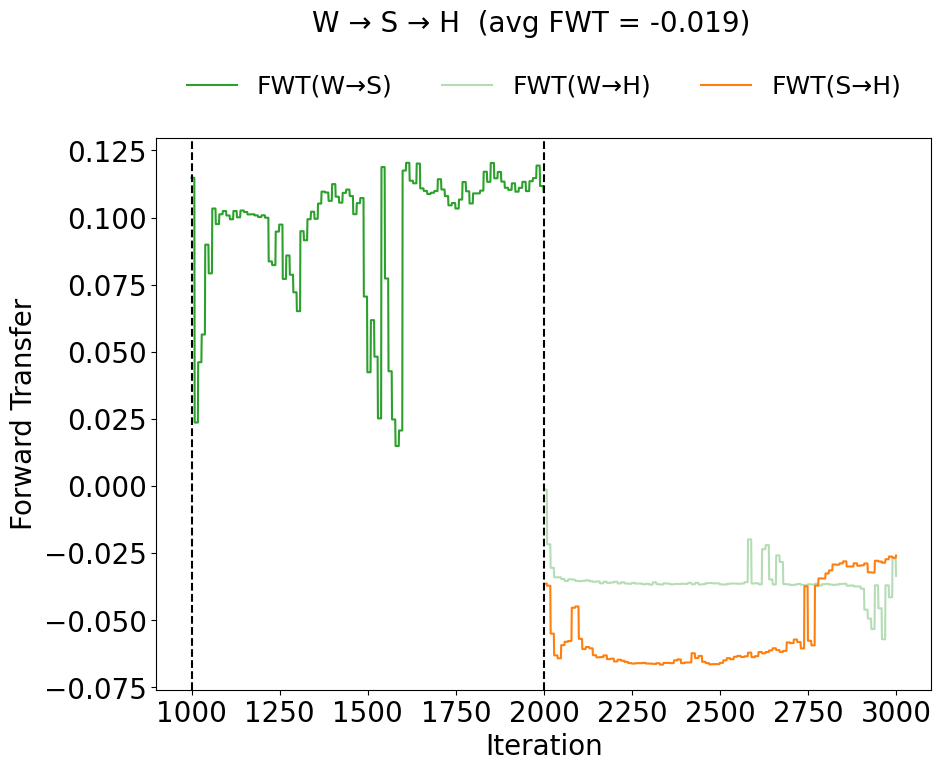}
\end{subfigure}

\caption{Forward transfer (FWT) during sequential continual learning for the shared output head policies. Each plot corresponds to a different task ordering across the three MuJoCo environments (\textit{Hopper-v5}, \textit{Swimmer-v5}, and \textit{Walker2d-v5}). Curves show the evolution of task-wise forward transfer (FWT) over training iterations, with vertical dashed lines marking task transitions. Normalization is done with respect to the maximum reward achieved in the corresponding single-task runs.}
\Description{Forward transfer (FWT) during sequential continual learning for the shared output head policies. Each plot corresponds to a different task ordering across the three MuJoCo environments (\textit{Hopper-v5}, \textit{Swimmer-v5}, and \textit{Walker2d-v5}). Curves show the evolution of task-wise forward transfer (FWT) over training iterations, with vertical dashed lines marking task transitions. Normalization is done with respect to the maximum reward achieved in the corresponding single-task runs.}
\label{fig:fwt_shared}
\end{figure*}

\begin{figure*}[p]
\centering
\setkeys{Gin}{height=0.23\textheight,keepaspectratio}

\begin{subfigure}{0.45\textwidth}
    \centering
    \includegraphics[width=\linewidth]{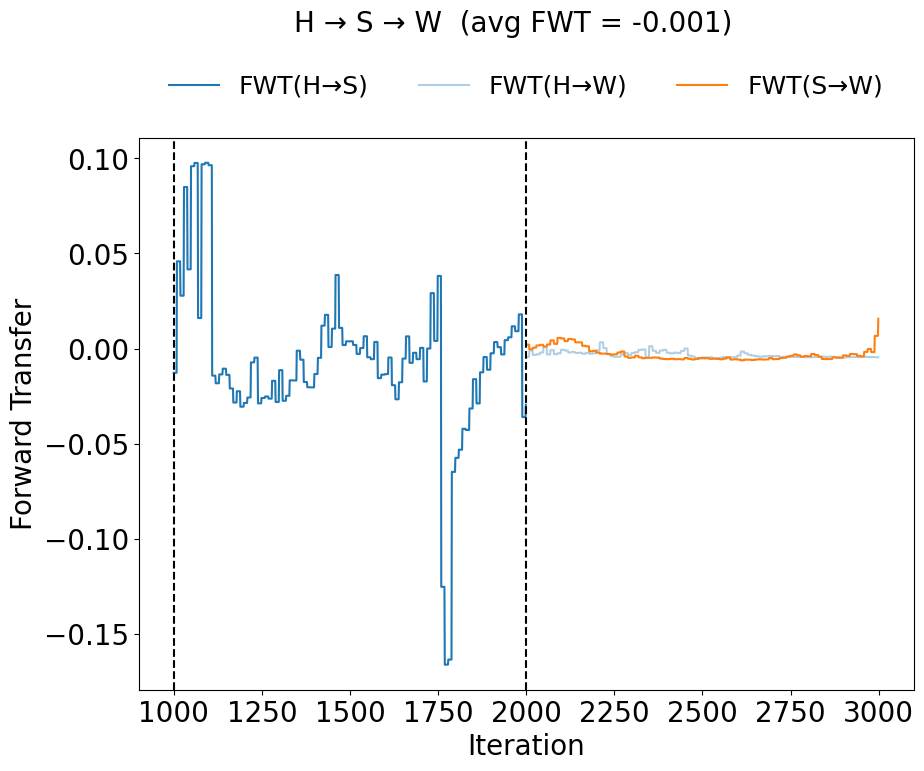}
    \includegraphics[width=\linewidth]{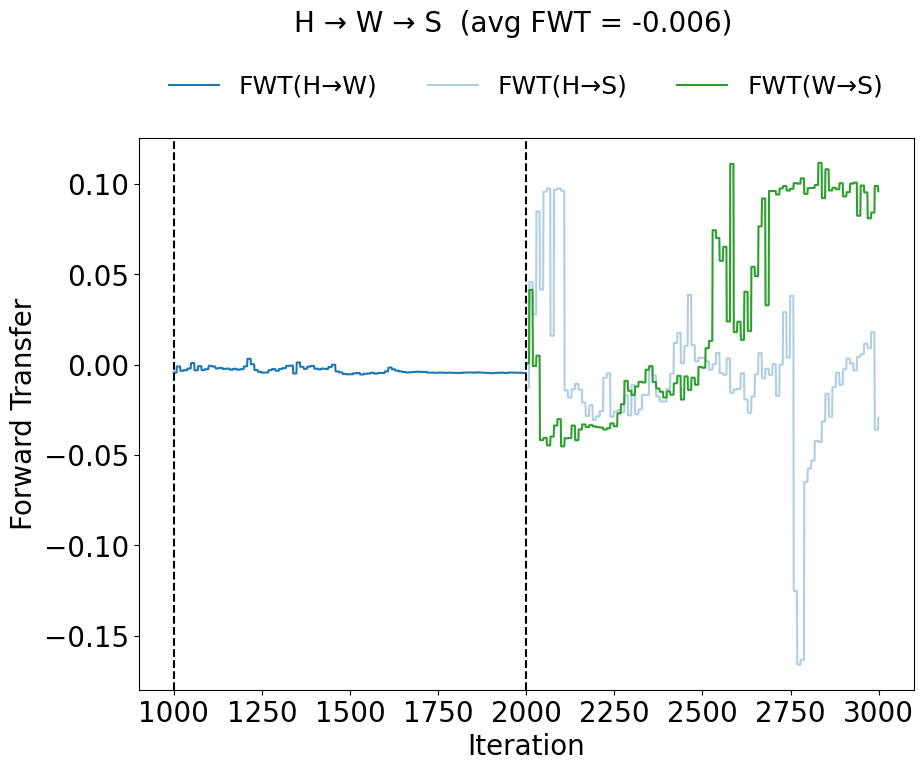}
    \includegraphics[width=\linewidth]{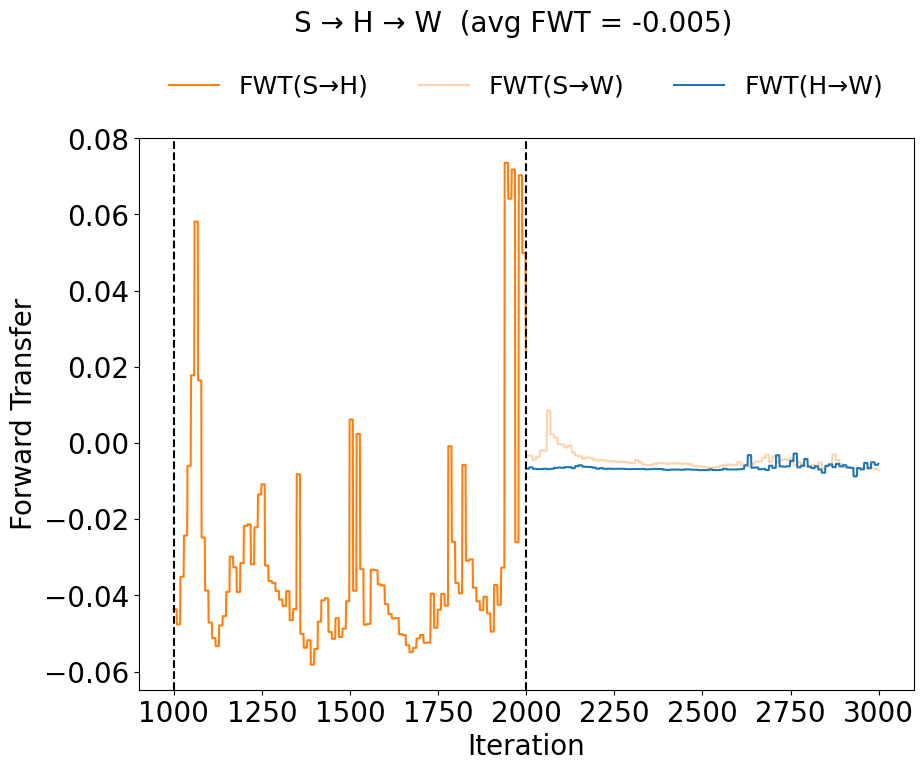}
\end{subfigure}\hfill
\begin{subfigure}{0.45\textwidth}
    \centering
    \includegraphics[width=\linewidth]{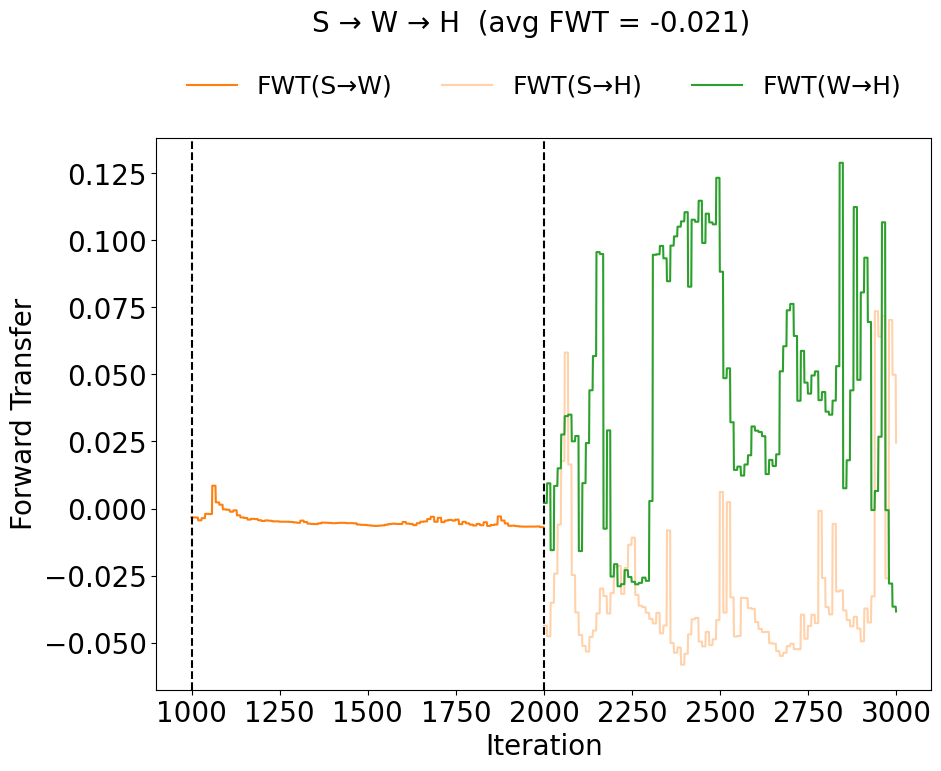}
    \includegraphics[width=\linewidth]{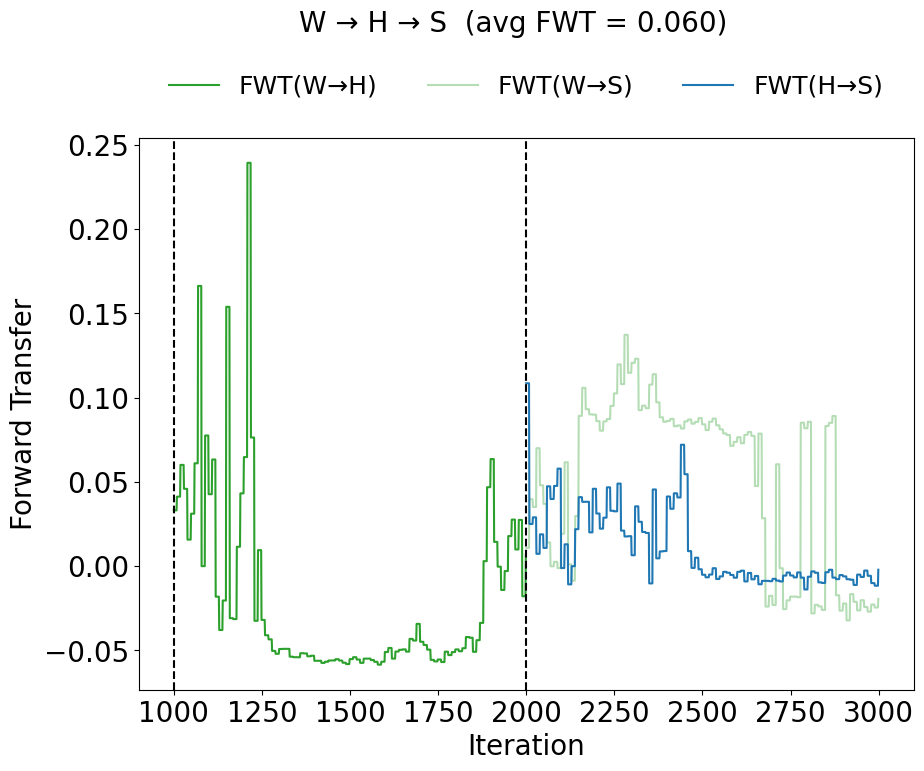}
    \includegraphics[width=\linewidth]{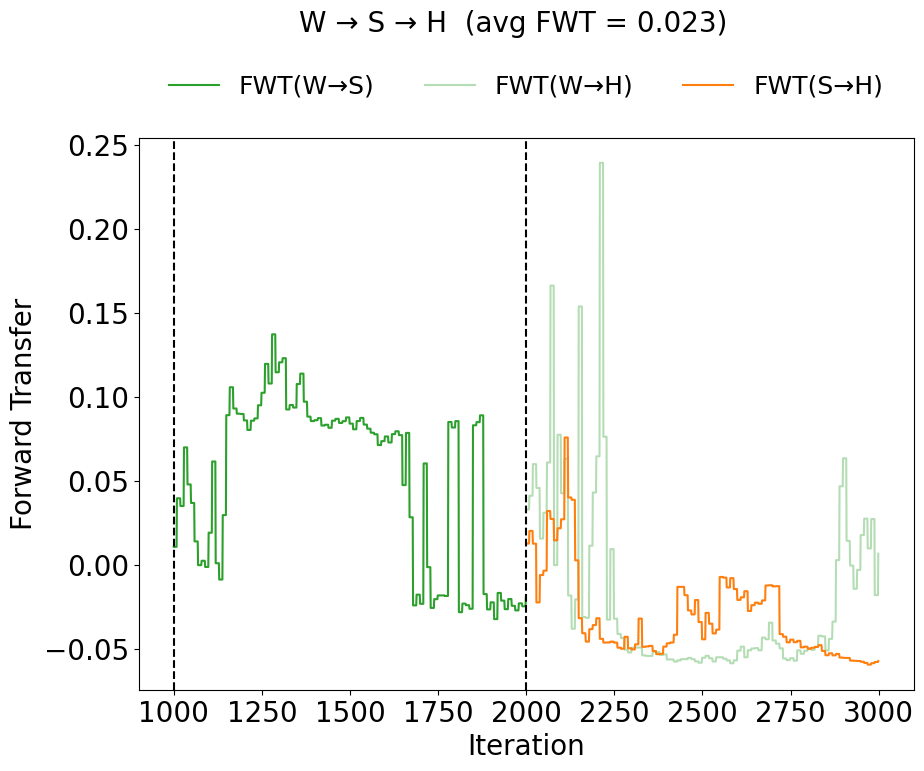}
\end{subfigure}

\caption{Forward transfer (FWT) during sequential continual learning for the distinct output head policies. Each plot corresponds to a different task ordering across the three MuJoCo environments (\textit{Hopper-v5}, \textit{Swimmer-v5}, and \textit{Walker2d-v5}). Curves show the evolution of task-wise forward transfer (FWT) over training iterations, with vertical dashed lines marking task transitions. Normalization is done with respect to the maximum reward achieved in the corresponding single-task runs.}
\Description{Forward transfer (FWT) during sequential continual learning for the distinct output head policies. Each plot corresponds to a different task ordering across the three MuJoCo environments (\textit{Hopper-v5}, \textit{Swimmer-v5}, and \textit{Walker2d-v5}). Curves show the evolution of task-wise forward transfer (FWT) over training iterations, with vertical dashed lines marking task transitions. Normalization is done with respect to the maximum reward achieved in the corresponding single-task runs.}
\label{fig:fwt_distinct}
\end{figure*}

Here, we report all six task permutations, providing both learning dynamics and extrema statistics. Learning curves for the shared-output agent (Figure \ref{fig:sequential_shared}) highlight a strong dependence on task order and a pronounced drop on previously learned tasks after each transition, i.e., substantial forgetting; the corresponding maxima/minima at each step are reported in Table \ref{tab:maxmin-rewards-shared}. Distinct output heads improve stability across permutations (Figure \ref{fig:sequential_distinct}), and their step-wise best/worst returns are summarized in Table \ref{tab:maxmin-rewards-distinct}. Adding replay further preserves earlier-task performance and reduces degradation after transitions (Figure \ref{fig:sequential_replay}); the associated extrema across all orders are reported in Table \ref{tab:maxmin-rewards-replay}, while Table \ref{tab:maxmin-rewards-hsw-replay} isolates the sequence \textit{Hopper} $\to$ \textit{Swimmer} $\to$ \textit{Walker2d} to quantify the effect of varying replay budgets.
Backward transfer results are consistent with these trends: shared-output policies exhibit mostly negative BWT (Figure \ref{fig:bwt_shared}), while replay yields less negative and occasionally positive BWT values (Figure \ref{fig:bwt_replay}); distinct-output policies show the most stable BWT profiles with reduced interference (Figure \ref{fig:bwt_distinct}). Finally, forward transfer is reported for shared and distinct architectures (Figures \ref{fig:fwt_shared} and \ref{fig:fwt_distinct}), where distinct outputs provide more consistent positive transfer across task orders.

\begin{table*}[p]
\centering
\scriptsize
\resizebox{\textwidth}{!}{%
\begin{tabular}{cc}
\begin{tabular}{lccc}
\multicolumn{4}{c}{\textbf{Hopper $\to$ Swimmer $\to$ Walker2d}} \\ \hline
Step & Hopper & Swimmer & Walker2d \\ \hline
1 & 1199.98 / 81.94 & -- & -- \\
2 & 1214.35 / 99.00 & 359.78 / 9.46 & -- \\
3 & 728.62 / 59.66 & 362.27 / 64.51 & 1093.85 / $-1.13$ \\
\end{tabular}
&
\begin{tabular}{lccc}
\multicolumn{4}{c}{\textbf{Hopper $\to$ Walker2d $\to$ Swimmer}} \\ \hline
Step & Hopper & Walker2d & Swimmer \\ \hline
1 & 1199.98 / 81.94 & -- & -- \\
2 & 1203.32 / 79.79 & 2102.04 / $-4.45$ & -- \\
3 & 94.87 / 46.85 & 2206.60 / 154.42 & 55.04 / $-22.34$ \\
\end{tabular}
\\[2em]
\begin{tabular}{lccc}
\multicolumn{4}{c}{\textbf{Swimmer $\to$ Walker2d $\to$ Hopper}} \\ \hline
Step & Swimmer & Walker2d & Hopper \\ \hline
1 & 359.67 / 28.87 & -- & -- \\
2 & 359.64 / 33.13 & 1556.08 / 7.86 & -- \\
3 & 340.80 / 35.19 & 1553.43 / 184.02 & 1102.77 / 50.22 \\
\end{tabular}
&
\begin{tabular}{lccc}
\multicolumn{4}{c}{\textbf{Swimmer $\to$ Hopper $\to$ Walker2d}} \\ \hline
Step & Swimmer & Hopper & Walker2d \\ \hline
1 & 359.67 / 28.87 & -- & -- \\
2 & 361.93 / 19.45 & 1115.78 / 4.24 & -- \\
3 & 345.75 / 94.56 & 1120.95 / 210.16 & 1809.03 / 13.61 \\
\end{tabular}
\\[2em]
\begin{tabular}{lccc}
\multicolumn{4}{c}{\textbf{Walker2d $\to$ Hopper $\to$ Swimmer}} \\ \hline
Step & Walker2d & Hopper & Swimmer \\ \hline
1 & 3416.02 / 58.71 & -- & -- \\
2 & 3468.34 / 649.40 & 1097.96 / 280.59 & -- \\
3 & 3314.69 / 341.66 & 1098.77 / 1084.51 & 110.58 / $-22.21$ \\
\end{tabular}
&
\begin{tabular}{lccc}
\multicolumn{4}{c}{\textbf{Walker2d $\to$ Swimmer $\to$ Hopper}} \\ \hline
Step & Walker2d & Swimmer & Hopper \\ \hline
1 & 3416.02 / 58.71 & -- & -- \\
2 & 3442.05 / 682.21 & 355.06 / 24.10 & -- \\
3 & 1901.46 / 58.80 & 362.14 / 45.63 & 1098.01 / 55.72 \\
\end{tabular}
\end{tabular}
} %
\caption{Maximum and minimum rewards obtained for each of the six task orders (shared-output). Each cell reports ``max / min''. ``--'' indicates the task had not yet been introduced.}
\Description{Maximum and minimum rewards obtained for each of the six task orders (shared-output). Each cell reports ``max / min''. ``--'' indicates the task had not yet been introduced.}
\label{tab:maxmin-rewards-shared}
\end{table*}

\begin{table*}[p]
\centering
\scriptsize
\resizebox{\textwidth}{!}{%
\begin{tabular}{cc}
\begin{tabular}{lccc}
\multicolumn{4}{c}{\textbf{Hopper $\to$ Swimmer $\to$ Walker2d}} \\ \hline
Step & Hopper & Swimmer & Walker2d \\ \hline
1 & 1199.98 / 81.94 & -- & -- \\
2 & 1235.77 / 1184.77 & 280.54 / 12.12 & -- \\
3 & 1304.45 / 1211.04 & 281.30 / 271.60 & 3545.12 / 20.89 \\
\end{tabular}
&
\begin{tabular}{lccc}
\multicolumn{4}{c}{\textbf{Hopper $\to$ Walker2d $\to$ Swimmer}} \\ \hline
Step & Hopper & Walker2d & Swimmer \\ \hline
1 & 1199.98 / 81.94 & -- & -- \\
2 & 1383.36 / 772.46 & 1857.66 / $-5.23$ & -- \\
3 & 1620.27 / 1319.83 & 1854.83 / 1331.11 & 159.58 / $-17.81$ \\
\end{tabular}
\\[2em]
\begin{tabular}{lccc}
\multicolumn{4}{c}{\textbf{Swimmer $\to$ Walker2d $\to$ Hopper}} \\ \hline
Step & Swimmer & Walker2d & Hopper \\ \hline
1 & 359.67 / 28.87 & -- & -- \\
2 & 362.73 / 273.03 & 1523.99 / 9.93 & -- \\
3 & 361.89 / 351.87 & 1677.78 / 1306.91 &
1110.39 / 15.06 \\
\end{tabular}
&
\begin{tabular}{lccc}
\multicolumn{4}{c}{\textbf{Swimmer $\to$ Hopper $\to$ Walker2d}} \\ \hline
Step & Swimmer & Hopper & Walker2d \\ \hline
1 & 359.67 / 28.87 & -- & -- \\
2 & 361.77 / 277.80 & 2144.73 / 69.35 & -- \\
3 & 362.69 / 317.26 & 1899.98 / 1662.79 & 1944.47 / 3.73 \\
\end{tabular}
\\[2em]
\begin{tabular}{lccc}
\multicolumn{4}{c}{\textbf{Walker2d $\to$ Hopper $\to$ Swimmer}} \\ \hline
Step & Walker2d & Hopper & Swimmer \\ \hline
1 & 3416.02 / 58.71 & -- & -- \\
2 & 3438.22 / 2924.85 & 1098.53 / 268.33 & -- \\
3 & 3395.50 / 3128.14 & 1098.36 / 1086.01 & 261.65 / 14.14 \\
\end{tabular}
&
\begin{tabular}{lccc}
\multicolumn{4}{c}{\textbf{Walker2d $\to$ Swimmer $\to$ Hopper}} \\ \hline
Step & Walker2d & Swimmer & Hopper \\ \hline
1 & 3416.02 / 58.71 & -- & -- \\
2 & 3470.82 / 3315.18 & 360.35 / 26.82 & -- \\
3 & 3420.46 / 3275.49 & 359.13 / 344.87 & 2694.75 / 38.00 \\
\end{tabular}
\end{tabular}
} %
\caption{Maximum and minimum rewards obtained with replay for each of the six task orders (replay). Each cell reports ``max / min''. ``--'' indicates the task had not yet been introduced.}
\Description{Maximum and minimum rewards obtained with replay for each of the six task orders (replay). Each cell reports ``max / min''. ``--'' indicates the task had not yet been introduced.}
\label{tab:maxmin-rewards-replay}
\end{table*}

\clearpage

\begin{table*}[p]
\centering
\scriptsize
\resizebox{\textwidth}{!}{%
\begin{tabular}{cc}
\begin{tabular}{lccc}
\multicolumn{4}{c}{\textbf{Hopper $\to$ Swimmer $\to$ Walker2d (replay = 12)}} \\ \hline
Step & Hopper & Swimmer & Walker2d \\ \hline
1 & 1199.98 / 81.94 & -- & -- \\
2 & 1208.42 / 1071.16 & 361.57 / 3.24 & -- \\
3 & 1109.83 / 170.40 & 362.20 / 45.81 & 1682.98 / $-3.60$ \\
\end{tabular}
&
\begin{tabular}{lccc}
\multicolumn{4}{c}{\textbf{Hopper $\to$ Swimmer $\to$ Walker2d (replay = 192)}} \\ \hline
Step & Hopper & Swimmer & Walker2d \\ \hline
1 & 1199.98 / 81.94 & -- & -- \\
2 & 1235.77 / 1184.77 & 280.54 / 12.12 & -- \\
3 & 1304.45 / 1211.04 & 281.30 / 271.60 & 3545.12 / 20.89 \\
\end{tabular}
\\[2em]
\multicolumn{2}{c}{
\begin{tabular}{lccc}
\multicolumn{4}{c}{\textbf{Hopper $\to$ Swimmer $\to$ Walker2d (replay = 288)}} \\ \hline
Step & Hopper & Swimmer & Walker2d \\ \hline
1 & 1199.98 / 81.94 & -- & -- \\
2 & 1246.31 / 1184.47 & 281.19 / 6.99 & -- \\
3 & 1261.61 / 1211.75 & 280.35 / 273.02 & 1972.13 / 0.82 \\
\end{tabular}
}
\end{tabular}
} %
\caption{Maximum and minimum unnormalized rewards for the sequence Hopper $\to$ Swimmer $\to$ Walker2d under three replay budgets. Each cell reports ``max / min''. ``--'' indicates the task had not yet been introduced.}
\Description{Maximum and minimum unnormalized rewards for the sequence Hopper $\to$ Swimmer $\to$ Walker2d under three replay budgets. Each cell reports ``max / min''. ``--'' indicates the task had not yet been introduced.}
\label{tab:maxmin-rewards-hsw-replay}
\end{table*}

\begin{table*}[p]
\centering
\scriptsize
\resizebox{\textwidth}{!}{%
\begin{tabular}{cc}
\begin{tabular}{lccc}
\multicolumn{4}{c}{\textbf{Hopper $\to$ Swimmer $\to$ Walker2d}} \\ \hline
Step & Hopper & Swimmer & Walker2d \\ \hline
1 & 1167.33 / 74.43 & -- & -- \\
2 & 1349.07 / 1155.45 & 362.37 / 42.92 & -- \\
3 & 1167.31 / 71.43 & 362.27 / 13.60 & 2008.55 / $-3.51$ \\
\end{tabular}
&
\begin{tabular}{lccc}
\multicolumn{4}{c}{\textbf{Hopper $\to$ Walker2d $\to$ Swimmer}} \\ \hline
Step & Hopper & Walker2d & Swimmer \\ \hline
1 & 1167.33 / 74.43 & -- & -- \\
2 & 1224.14 / 229.38 & 2177.69 / 11.42 & -- \\
3 & 1193.51 / 1139.52 & 2230.53 / 1822.74 & 362.03 / 39.52 \\
\end{tabular}
\\[2em]
\begin{tabular}{lccc}
\multicolumn{4}{c}{\textbf{Swimmer $\to$ Walker2d $\to$ Hopper}} \\ \hline
Step & Swimmer & Walker2d & Hopper \\ \hline
1 & 362.21 / $-11.88$ & -- & -- \\
2 & 361.95 / 65.78 & 2370.11 / 11.61 & -- \\
3 & 359.84 / 45.74 & 2564.95 / 229.29 & 1184.28 / 59.16 \\
\end{tabular}
&
\begin{tabular}{lccc}
\multicolumn{4}{c}{\textbf{Swimmer $\to$ Hopper $\to$ Walker2d}} \\ \hline
Step & Swimmer & Hopper & Walker2d \\ \hline
1 & 362.21 / $-11.88$ & -- & -- \\
2 & 361.70 / 29.28 & 2455.61 / 80.94 & -- \\
3 & 360.21 / 42.21 & 2495.32 / 237.56 & 2219.82 / 2.72 \\
\end{tabular}
\\[2em]
\begin{tabular}{lccc}
\multicolumn{4}{c}{\textbf{Walker2d $\to$ Hopper $\to$ Swimmer}} \\ \hline
Step & Walker2d & Hopper & Swimmer \\ \hline
1 & 3743.99 / 23.98 & -- & -- \\
2 & 3871.13 / 205.88 & 1107.09 / 19.05 & -- \\
3 & 3770.01 / 2885.06 & 1101.34 / 1097.26 & 359.76 / 28.92 \\
\end{tabular}
&
\begin{tabular}{lccc}
\multicolumn{4}{c}{\textbf{Walker2d $\to$ Swimmer $\to$ Hopper}} \\ \hline
Step & Walker2d & Swimmer & Hopper \\ \hline
1 & 3743.99 / 23.98 & -- & -- \\
2 & 3935.82 / 3399.81 & 359.95 / 19.13 & -- \\
3 & 3720.81 / 244.34 & 362.26 / 33.22 & 1765.72 / 11.26 \\
\end{tabular}
\end{tabular}
} %
\caption{Maximum and minimum unnormalized rewards obtained for each of the six task orders (distinct-output). Each cell reports ``max / min''. ``--'' indicates the task had not yet been introduced.}
\Description{Maximum and minimum unnormalized rewards obtained for each of the six task orders (distinct-output). Each cell reports ``max / min''. ``--'' indicates the task had not yet been introduced.}
\label{tab:maxmin-rewards-distinct}
\end{table*}

\clearpage

\end{document}